\documentclass[preprint,5p,times,twocolumn]{elsarticle}

\usepackage{amsmath}
\usepackage{booktabs}

\usepackage{subfigure}
\usepackage{caption}
\usepackage{pythontex}
\usepackage{algorithm}
\usepackage{algpseudocode}
\usepackage{siunitx}
\usepackage{amssymb}
\usepackage{multirow}
\usepackage{adjustbox}
\usepackage{array}
\newcolumntype{C}[1]{>{\centering\arraybackslash}p{#1}}

\renewcommand{\algorithmiccomment}[1]{\bgroup\hfill//~#1\egroup}

\makeatletter
\renewcommand{\ALG@name}{ALGORITHM}
\makeatother

\usepackage{wrapfig}
\usepackage{graphicx}
\DeclareGraphicsExtensions{.pdf,.png,.jpg}
\usepackage{url}
\usepackage[breaklinks]{hyperref}
\usepackage{xurl}

\usepackage{subcaption}
\usepackage{multicol}
\usepackage[inline]{enumitem}
\usepackage{multirow}
\usepackage{listings}
\usepackage{todonotes}
\definecolor{keywords}{RGB}{255,0,90}
\definecolor{comments}{RGB}{0,0,113}
\definecolor{red}{RGB}{160,0,0}
\definecolor{green}{RGB}{0,150,0}

\usepackage[most]{tcolorbox}

\newtcolorbox{qblock}{
  colback=yellow!30,     
  colframe=black,        
  fonttitle=\bfseries,   
  coltitle=black,        
  rounded corners,       
  boxrule=1pt,           
  arc=5mm,               
  boxsep=5pt,            
  left=5mm,              
  right=5mm,             
  top=2mm,               
  bottom=2mm             
}

\newtcolorbox{ablock}{
  colback=blue!20,     
  colframe=black,        
  fonttitle=\bfseries,   
  coltitle=black,        
  rounded corners,       
  boxrule=1pt,           
  arc=5mm,               
  boxsep=5pt,            
  left=5mm,              
  right=5mm,             
  top=2mm,               
  bottom=2mm             
}

\journal{Journal of Systems and Software}

\begin{document}

\begin{frontmatter}

\title{Improving the Reproducibility of Deep Learning Software: An Initial Investigation through a Case Study Analysis}

\author[Purdue]{Nikita Ravi}
\author[NVIDIA]{Abhinav Goel}
\author[Purdue]{James C. Davis}
\author[Loyola]{George K. Thiruvathukal}

\affiliation[Purdue]{
            organization={Purdue University},
            city={West Lafayette},
            postcode={47906}, 
            state={IN},
            country={U.S.A}}

\affiliation[NVIDIA]{
            organization={NVIDIA},
            city={Santa Clara},
            postcode={95051}, 
            state={CA},
            country={U.S.A}}

\affiliation[Loyola]{
            organization={Loyola University Chicago},
            city={Chicago},
            postcode={60660}, 
            state={IL},
            country={U.S.A}}


\begin{abstract}

The field of deep learning has witnessed significant breakthroughs, spanning various applications,  and fundamentally transforming current software capabilities. 
However, alongside these advancements, there have been increasing concerns about reproducing the results of these deep learning methods. 
This is significant because reproducibility is the foundation of reliability and validity in software development, particularly in the rapidly evolving domain of deep learning.
The difficulty of reproducibility may arise due to several reasons, including having differences from the original execution environment, missing or incompatible software libraries, proprietary data and source code, lack of transparency in the data-processing and training pipeline, and the stochastic nature in some software. A study conducted by the Nature journal reveals that more than 70\% of researchers failed to reproduce other researcher's experiments and over 50\% failed to reproduce their own experiments. Given the critical role that deep learning plays in many software applications, irreproducibility poses significant challenges for researchers and practitioners. 
To address these concerns, this paper presents a systematic approach at analyzing and improving the reproducibility of deep learning models by demonstrating these guidelines using a case study. We illustrate the patterns and anti-patterns involved with these guidelines for improving the reproducibility of deep learning models. These guidelines encompass establishing a methodology to replicate the original software environment, implementing end-to-end training and testing algorithms, disclosing architectural designs, and enhancing transparency in data processing and training pipelines. We also conduct a sensitivity analysis to understand the model's performance across diverse conditions.
By implementing these strategies, we aim to bridge the gap between research and practice, so that innovations in deep learning can be effectively reproduced and deployed within software. 
\end{abstract}


\end{frontmatter}

\section{Introduction and Motivation}
Deep learning (DL) has emerged as a cornerstone in modern technology with applications in healthcare, autonomous vehicles, natural language processing,  computer vision, and many more.
Once researchers showcase the potential of a deep learning approach in addressing a problem, organizations may incorporate this method into their software products \cite{CVreengineering}. 
Yet, the journey from research to practice, solving real-world problems using deep learning software presents complex challenges due to the lack of reproducibility \cite{maskformer}. 
Reproducing the performance of deep learning models is a foundational aspect of scientific integrity. 
It ensures that findings are reliable, verifiable, and applicable across various environments, even in different execution environments with variations in hardware and software \cite{missData, Lemay2022-am}. 
Reproducibility enables the scientific community and industries to adopt and implement deep learning software with confidence. 

\begin{table*}[t] 
\centering
\small
 \begin{tabular}{p{0.17\textwidth}C{1.2cm}C{1.2cm}C{1.2cm}C{1.2cm}C{1.2cm}C{1.2cm}C{1.2cm}C{1.2cm}} 
 \toprule
 \textbf{Paper} & \textbf{ES} & \textbf{HW} & \textbf{Datasets} & \textbf{Train}  &\textbf{SA}& \textbf{Test} & \textbf{Doc.} & \textbf{CS}\\ \midrule
Artrith et al. \cite{ml_chem} & $\times$ & $\times$ & \checkmark & $\times$  &$\times$& $\times$ & $\times$ & $\times$\\ 
Chen et al. \cite{chen2022} & $\times$ & \checkmark & $\times$ & $\times$  &$\times$& $\times$ & $\times$ & $\times$ \\ 
Haibe-Kains et al. \cite{natureTransp} & \checkmark & $\times$ & \checkmark & \checkmark  &$\times$& $\times$ & $\times$ & \checkmark \\ 
Isdahl et al. \cite{outofbox} & \checkmark & \checkmark & \checkmark & \checkmark  &$\times$& $\times$ & $\times$ & $\times$ \\ 
Pineau et al. \cite{pineau2020} & $\times$ & $\times$ & \checkmark & \checkmark  &$\times$& $\times$ & $\times$ & $\times$ \\ 
Semmelrock et al. \cite{semmelrock2023reproducibility} & \checkmark & \checkmark & \checkmark & $\times$  &$\times$& $\times$ & $\times$ & $\times$ \\ \midrule
This paper & \checkmark & \checkmark & \checkmark & \checkmark  &\checkmark& \checkmark & \checkmark & \checkmark\\ \bottomrule
 \end{tabular}
 \caption{Comparing Contributions of Reproducibility for Deep Learning Research. We check whether each paper mentions the significance of providing what is mentioned in each column. \textbf{ES}: Providing instructions for Environment Setup. \textbf{HW}: Providing description of Hardware. \textbf{Datasets}: Explaining the need to provide the data processing steps. \textbf{Train} and \textbf{Test}: Explaining the significance of disclosing the end to end training and testing process. \textbf{SA}: Explaining the significance of conducting a sensitivity analysis. \textbf{Doc}: Explaining the importance of documenting deep learning software. \textbf{CS}: Using a case study to provide meaningful examples.
 }
 \label{tab:otherPapers}
\end{table*}

A 2016 survey by Nature \cite{nature2016} revealed that out of the 1,572 researchers surveyed, 52\% agreed that there is a significant ``crisis" of reproducibility. More than 70\% researchers failed to reproduce other researchers' experiments and over 50\% failed to reproduce their own experiments. Furthermore, the ease with which machine learning is promoted as a tool that can be quickly applied by researchers across various disciplines has led to concerns about a ``brewing reproducibility crisis", according to researchers at Princeton University \cite{natureGibney}. 

The software engineering task of reusing, reproducing, and adapting cutting-edge deep learning approaches is challenging due to many reasons, such as lack of source code or data, unclear documentation, different execution environments, or stochastic nature of many machine learning models~\cite{ml_chem, chen2022, pineau2020, semmelrock2023reproducibility, Gundersen_Kjensmo_2018, pmlr-v97-bouthillier19a}.

\begin{figure}[h]
\centering
\includegraphics[width=0.50\textwidth]{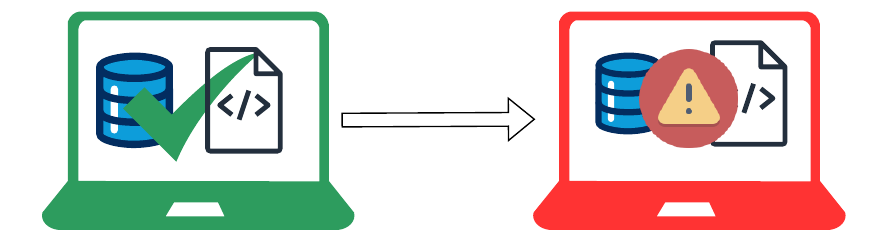}
\caption{Even with the provided source code and datasets, reproducibility challenges in deep learning still persist \cite{pineau2020}.}
\label{fig:reproDef}
\end{figure}

To tackle the challenges of reproducibility in deep learning software, this paper outlines a comprehensive set of strategies designed to enhance the reproducibility of these systems through the illustration of a case study known as the \textbf{Tr}ee-Based \textbf{U}nidirectional \textbf{N}eural Networ\textbf{k} (TRUNK) \cite{goel_modular_2020, TRUNK_goel}. We chose TRUNK as our case study due to its complexity and robustness as a deep learning method, which also aligns well with our hardware capabilities as we evaluate the reproducibility guidelines.

These guidelines include (1) establishing a robust methodology to set up the software environment, (2) implementing end-to-end training and testing algorithms, (3) disclosing architectural designs, (4) enhancing the transparency in the data processing and training pipelines, and finally (5) conducting a sensitivity analysis to gain a deeper understanding of the model's behavior across diverse conditions. Even though the inherent stochastic nature of some deep learning models presents unique challenges, the guidelines aim to equip researchers with effective strategies to improve the reproducibility of training deep learning models, despite this obstacle. 

There have been many studies that have delved into the concept of reproducibility in deep learning \cite{ml_chem, chen2022, natureTransp, pineau2020, semmelrock2023reproducibility, Gundersen_Kjensmo_2018, pmlr-v97-bouthillier19a}. These studies propose various guidelines and rules to enhance reproducibility. However, there have not been enough case studies conducted to illustrate the application of these guidelines. The approach of using case studies is important for understanding the complexities involved and identifying any shortcomings in the current recommendations.

Table \ref{tab:otherPapers} outlines the contributions made in this paper by comparing the guidelines and solutions provided by other reproducibility-focused papers. The paper is organized as follows:
\begin{itemize}
    \item \S\ref{sec:reproDef} defines reproducibility and the obstacles that hinder the reproducibility of deep learning models.
    \item \S\ref{sec:currGuide} introduces the current recommended guidelines for improving the reproducibility of deep learning software.
    \item \S\ref{sec:LPCV}-\S\ref{sec:TRUNK} justifies the reason for using the TRUNK model as our case study
    \item \S\ref{sec:experiments} presents the experiments conducted for analyzing the reproducibility of our case study and their respective results.
    \item \S\ref{sec:guidelines} introduces our strategies for improving the reproducibility of training deep learning models, based on the lessons learnt from our experiments.
\end{itemize}

The source code, pre-trained models, and guidelines are available on \href{https://github.com/nikkiravi/TRUNK-Reproducibility/}{GitHub}. 
\section{Background and Related Works}\label{sec:reproBack}
\subsection{Reproducibility of Deep Learning Software} \label{sec:reproDef}
Software engineering involves reproducing, reusing, understanding, or improving existing implementations \cite{CVreengineering, davis2024reusing}. 
In contrast to traditional software engineering, the reproducibility of deep learning software is a relatively new focus area \cite{davis2024reusing, frakesReuse, kruegerReuse, 10.1109/ICSE-SEIP.2019.00042, 9266043}. 
Pineau et al. defines reproducibility in deep learning as the process of re-doing an experiment using the same data and analytical tools to derive the same conclusions \cite{pineau2020, semmelrock2023reproducibility}. 
As the field of deep learning evolves, emphasizing reproducibility will be essential to validate and build upon prior work, fostering a more robust and reliable software development environment.

\begin{table}[h]
\centering
\resizebox{0.47\textwidth}{!}{%
\begin{tabular}{ll|cc|}
\cline{3-4}
\multirow{2}{*}{} &  & \multicolumn{2}{c|}{\textbf{Data}} \\ \cline{3-4} 
    &   & \multicolumn{1}{c|}{\textbf{Same}} & \textbf{Different} \\ \hline
\multicolumn{1}{|c|}{\multirow{2}{*}{\textbf{Code and Analysis}}} & \multicolumn{1}{c|}{\textbf{Same}}  & \multicolumn{1}{c|}{Reproducible} & Replicable \\ \cline{2-4} 
\multicolumn{1}{|c|}{}  & \multicolumn{1}{c|}{\textbf{Different}} & \multicolumn{1}{c|}{Robust} & Generalizable \\ \hline
\end{tabular}%
}
\caption{Reproducibility for Deep Learning Research is classified by whether similar results were obtained using the same code and data \cite{pineau2020}}
\label{tab:reproDef}
\end{table}

Deep learning represents a fundamental shift in software development, heralding the era of Software 2.0. Unlike the traditional Software 1.0, which uses explicit, human-written instructions in languages like Python and C++, Software 2.0 operates with abstract representations such as neural network weights. These weights, often numbering in the millions, are not manually coded by humans due to their complexity and volume \cite{SWEvsAI}. 
This transition in the software engineering task of reproducing and adapting DL software \cite{davis2024reusing} is challenging for reasons including:

\begin{enumerate}
    \item \textbf{Differences in Execution Environments}: Studies have shown that having a different software environment from the original implementation can contribute to a situation where deep learning software is not easily reproducible \cite{semmelrock2023reproducibility}. For example, Pouchard et al. \cite{MLinMSE} were unable to reproduce the performance of a multi-layer perceptron that was originally implemented in TensorFlow, using PyTorch and vice-versa despite using the same random seed and the same datasets.
    Many DL processes, unlike traditional software \cite{SWEvsAI}, depend heavily on pseudorandom sequences, making Pseudorandom Number Generators (PRNGs) indispensable \cite{chen2022, seed, antunes2024reproducibility}. 
    Antunes et al. revealed that different DL libraries use diverse PRNGs and handle their initialization and state management differently, posing challenges to reproducibility \cite{antunes2024reproducibility}.
    
    The current recommendations in the literature advocate the use of package managers \cite{natureTransp, semmelrock2023reproducibility} such as \texttt{conda} \cite{conda}, containers like Docker \cite{docker, dockerRepro}, and virtualization systems like Code Ocean \cite{codeOcean}, Gigantum \cite{Gigantum}, and Colaboratory \cite{Colaboratory}.  
    
    \item \textbf{Missing Data and Code}: A study conducted by the Norwegian University of Science and Technology (NTNU) \cite{missData} revealed that in a survey of 400 algorithms that were presented in papers at two top AI conferences, only 6\% shared the algorithm's code and a third shared their data. There are many reasons for the reluctance of sharing these materials, including having sensitive data/code or the increasing pressure for researchers to publish quickly. Such pressure often gives researchers little time to polish their code and decrease their willingness to release their code. 

    An analysis conducted by Haibe-Kains et al. \cite{natureTransp} revealed that researchers were hesitant to release the code used for training the models. The researchers claimed they had a ``large number of dependencies on internal tooling, infrastructure and hardware''  and therefore it was not possible to release the source code. In response to this belief, the study compiled a list of platforms that enable the sharing of code, software dependencies, and models. For example, the study suggested GitHub to share source code, the use of conda for software dependencies, TensorFlow Hub for the release of deep learning models, and the utilization of deep learning frameworks like PyTorch \cite{natureTransp}. Similarily, Isdahl et al. \cite{outofbox} presented a heatmap showing which software platforms have the necessary features to release code and data to facilitate for reproducible deep learning research. 
    
    \item \textbf{Randomness in the Software}: Randomness is essential in the process of training a deep learning model as it is involved in batch ordering, data shuffling, and weight initialization \cite{chen2022, seed, antunes2024reproducibility}. Randomness is one of the reasons why reproducibility in deep learning research cannot easily be achieved. Even if researchers provide both the code and the dataset, the random numbers generated throughout the training of a DL model can vary and lead to irreproducibility \cite{chen2022, hardwareDiff}. 
    Notably, Pham et al. \cite{hardwareDiff} demonstrated that despite standardizing the dependencies, the hardware, the seed, the datasets, and the source code, the accuracy of the deep learning model ranged from 8.6\% to 99.0\% due to the inherent randomness in the software.
    
    \item \textbf{Non-Determinism in the Hardware}: Training DL models typically requires intensive computing resources. Since GPUs have the ability to process multiple floating point operations in parallel, they are often used for DL training. However, executing floating point calculation in parallel \cite{gemmNVIDIA, reproTorch} becomes a source of non-determinism because the results of these operations are sensitive to the computation orders due to rounding errors \cite{goldberg}. 
\end{enumerate}


\subsection{The Current Guidelines for Improving Reproducibility of Deep Learning Software} \label{sec:currGuide}
Current guidelines proposed by various academic papers \cite{ml_chem, chen2022, natureTransp, outofbox, pineau2020, semmelrock2023reproducibility} highlight essential practices for improving the reproducibility of deep learning software. A compilation of these best practices are as follows:

\begin{tcolorbox} [width=\linewidth, top=1pt, bottom=1pt, left=2pt, right=2pt, title=Compilation of current guidelines for improving the reproducibility of DL software.]
    \begin{itemize}
        \item Provide resources to set up the software environment \cite{outofbox, pineau2020, semmelrock2023reproducibility, johnson2020datascience}
        \item Document the hardware used \cite{ml_chem, pineau2020}
        \item Initialize a seed \cite{chen2022, semmelrock2023reproducibility, seed} to limit the randomness in the software
        \item Disclose the data processing and training regime used \cite{ml_chem, natureTransp, torchDistill}
        \item Release the source code that would reproduce the paper onto public platforms \cite{ml_chem, natureTransp, outofbox, pineau2020, semmelrock2023reproducibility}
        \item Provide instructions on how to execute the source code \cite{johnson2020datascience}
        \item Provide access to the pre-trained weights \cite{johnson2020datascience}
        \item Release proper documentation alongside the DL software \cite{ml_chem, pineau2020, Gundersen_Kjensmo_2018, johnson2020datascience}
    \end{itemize}
\end{tcolorbox}



\section{Research Questions and Methodology} \label{sec:methods}
In this section, we outline the primary research questions driving our study for improving the reproducibility of deep learning software and detail the methodology employed to address these questions. 

\subsection{Research Questions}
To understand and analyze the process of enhancing the reproducibility of deep learning software, in this work we address the following research questions:

\begin{itemize}
    \item \textbf{RQ1:} How effective are the current recommended guidelines for improving the reproducibility of deep learning software?
    \item \textbf{RQ2:} How can the current guidelines for improving the reproducibility of deep learning software be extended or expanded?
\end{itemize}

\begin{figure*}[h]
\centering
\includegraphics[width=0.60\textwidth]{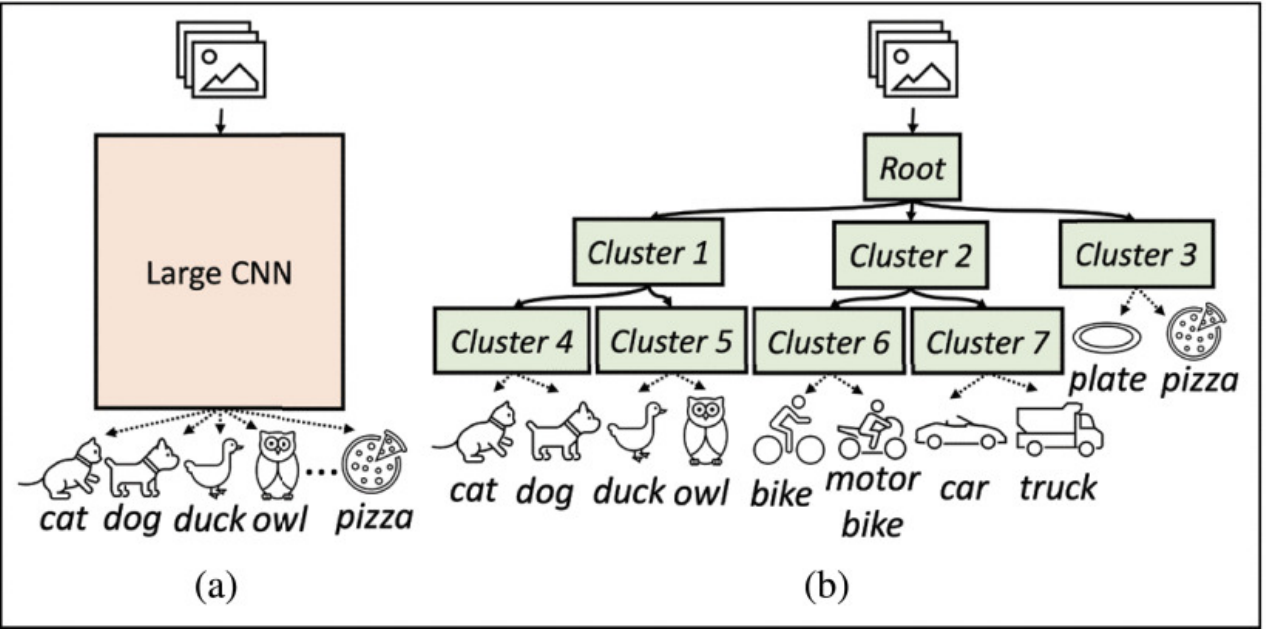}
\caption{(a) Monolithic Architectures vs (b) TRUNK \cite{TRUNK_goel}. We will use TRUNK, a type of hierarchical neural network, as a case study to demonstrate our guidelines for reproducibility.
} 
\label{fig:TRUNKvsMonolithic}
\end{figure*} 

\subsection{Choosing an Efficient Deep Learning Model as a Case Study} \label{sec:LPCV}
Through the use of a case study, we will analyze and assess the robustness of the current recommended practices for improving the reproducibiliy of a DL model. This approach allows for an in-depth examination of the successes and shortcomings of these guidelines in enhancing the reproducibility of DL software. We will present the patterns and anti-patterns observed in following these guidelines. Additionally, we will explore the potential for extending the current guidelines, particularly in areas where they have demonstrated shortcomings while reproducing our case study.

To understand why we choose the TRUNK model as our case study, it is essential to consider the nature and evolution of deep learning systems. Over the past decade, deep learning models have exponentially grown in size—from $10^8$ trainable parameters in AlexNet \cite{alexnet} to an astonishing $10^{13}$ parameters in GPT-4 \cite{gpt4}. Such large-scale models require significant resources; for instance, training GPT-4 involved over 10,000 NVIDIA A100 GPUs \cite{gpt4GPUs}. Given the vast resource requirements of contemporary models, our research opts for a more feasible approach, focusing on a model that aligns with the computational resources available to us to test our reproducibility guidelines.

Furthermore, deep learning has notably transformed computer vision software, drastically enhancing how machines perceive and analyze visual data. These advancements range from detecting whether individuals are wearing face masks \cite{faceMasks}, distinguishing different body poses \cite{bodyPoses}, to using semantic segmentation for autonomous vehicles to distinguish the features of their environment \cite{cityscape}. These critical advancements in technology motivates our decision to select a computer vision model that not only meets efficiency standards but also stands at the forefront of technological innovation in deep learning. One example of an efficient deep learning model for computer vision is the Tree-Based Unidirectional Neural Network (TRUNK) \cite{goel_modular_2020, TRUNK_goel}, which we have selected for our case study.

\subsection{Rationale for Choosing TRUNK as our Case Study for the Reproducibility Analysis} \label{sec:TRUNK}
Monolithic networks use a single DNN to identify every feature associated with all the categories to make a decision (Figure \ref{fig:TRUNKvsMonolithic}(a)). On the other hand, hierarchical networks, like that of TRUNK, are a collection of multiple shallow DNNs in the form of a tree and the leaf nodes represent each individual category of a particular dataset (Figure \ref{fig:TRUNKvsMonolithic}(b)). This hierarchical structure enables TRUNK to limit the number of redundant floating point operations that occur in order to classify a particular image \cite{goel_modular_2020}, therefore making it an efficient DL method for computer vision. 

\begin{figure*}[h]
\centering
\includegraphics[width=0.9\textwidth]{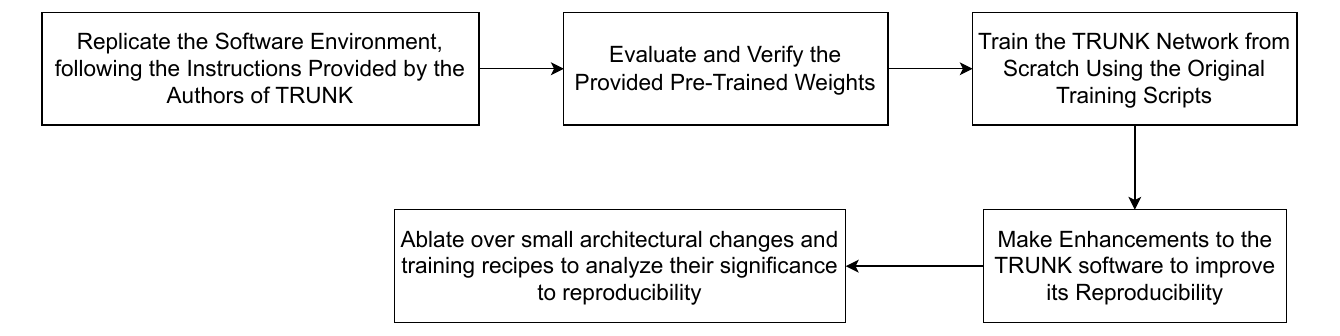}
\caption{Experiment Methodology used to Test the Reproducibility of TRUNK} 
\label{fig:reproExp}
\end{figure*} 

\begin{figure*}[h]
\begin{center}
\centering
\subfigure[]{\includegraphics[width=0.80\textwidth, height=3cm]{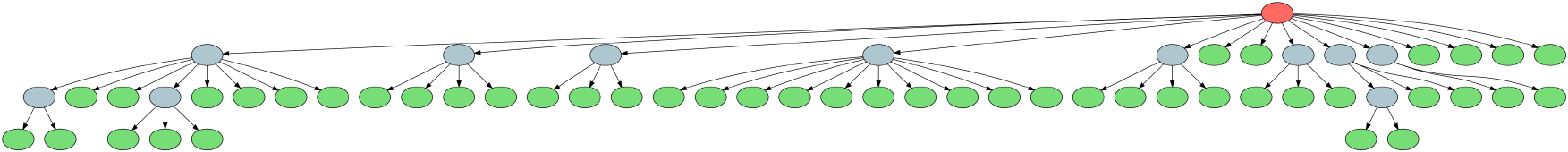}
}
\subfigure[]{\includegraphics[width=0.40\textwidth]{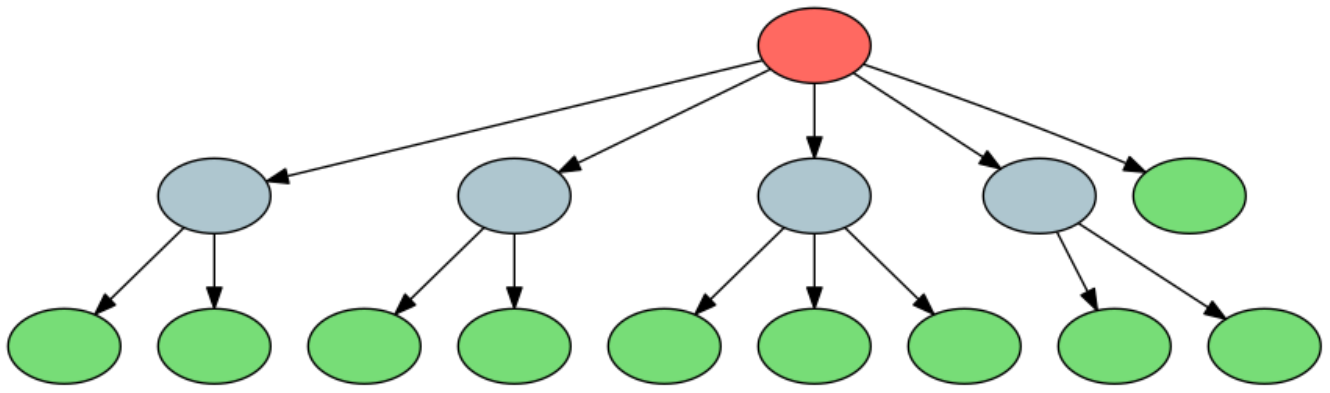}
}
\subfigure[]{\includegraphics[width=0.40\textwidth]{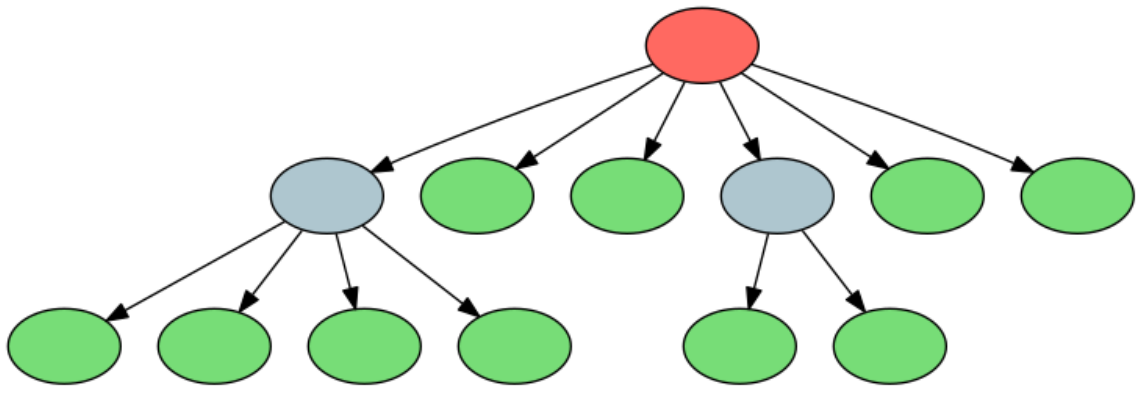}
}
\end{center}
\caption{We choose TRUNK for our case study analysis due to its complexity of the network architecture design varying by dataset (a) EMNIST (b) CIFAR-10 (c) SVHN. The red node is the root node of the tree, the gray nodes are the supergroups, and the green nodes are the leaf nodes.}
\label{fig:treeDataset}
\end{figure*}

TRUNK stands out among efficient hierarchical neural networks because it is one of the few \cite{goel_modular_2020, TRUNK_goel, bcnn, hdcnn, roy_tree-cnn:_2018} with an official GitHub repository that is entirely implemented in PyTorch. 
We preferred TRUNK over simpler, monolithic networks like VGG-16 \cite{simonyan_very_2014} or AlexNet \cite{alexnet} (Figure \ref{fig:TRUNKvsMonolithic}(a)) because of the added complexity it offers. 
Traditional DL models feature a consistent architecture which we train in a single pass. In contrast, TRUNK's architecture adapts based the dataset, as shown in Figure \ref{fig:treeDataset}, due to the visual similarity criteria \cite{goel_modular_2020, TRUNK_goel}. This tree structure of TRUNK is why it requires individual training for each node.
This unique approach not only tests the robustness of the reproducibility guidelines under complex training scenarios but also challenges our current computing resources. By focusing on TRUNK, we aim to rigorously assess the effectiveness of the guidelines in a demanding yet controlled environment, ensuring they can manage varying complexities in training deep learning models.

\subsection{Experiment Methodology} \label{sec:expSetUp}
The source code relevant to this study is publicly available for review on GitHub at \url{https://github.com/nikkiravi/TRUNK-Reproducibility/}. The \textit{LPCV Background} folder in the GitHub repository contains a Juypter notebook demonstrating the computational requirements of a basic CNN. The \textit{TRUNK} folder has the source code to train and conduct inference on the TRUNK architecture for the EMNIST \cite{cohen_emnist_2017}, CIFAR-10 \cite{krizhevsky_cifar}, and SVHN \cite{netzer2011reading} datasets. 

The experimental methodology to assess the robustness of the current reproducibility guidelines through the use of a case study is outlined in Figure \ref{fig:reproExp}. Initially, we will attempt to replicate the software environment used by the developers of TRUNK and explore ways to enhance the current methodology. 
Subsequently, we will assess the importance of providing pre-trained weights for enhancing reproducibility. This will be accomplished by verifying the results reported in the TRUNK paper and comparing them with those obtained from reproducing the training process of TRUNK.
Throughout this process, we will introduce improvements to the TRUNK software to bolster its reproducibility. Additionally, we will explore the impact of minor architectural modifications and variations in training recipes on the reproducibility of training deep learning models.

As we train TRUNK, we verify its reproducibility by comparing the accuracy of the pre-trained weights with that of our results. This comparison is conducted using a test dataset to provide an unbiased evaluation of classification accuracy. Additionally, we utilize a Python wrapper function to assess the total time required for conducting inference across the entire test dataset. The number of floating-point operations (FLOPs) executed across the network is measured using Python's \texttt{thop} library. For our reproducibility experiments, we employ Python version 3.9.18 and NVIDIA A100 GPUs.

The hyperparameters (i.e. batch size, epochs, learning rate, optimizer, learning rate scheduler, augmentations, and etc.) varied by dataset. We use the same hyperparameters provided by the authors of TRUNK for EMNIST \cite{cohen_emnist_2017} and SVHN \cite{netzer2011reading}. Since CIFAR-10 \cite{krizhevsky_cifar} lacked specific hyperparameter details, we applied the hyperparameters used for training SVHN to CIFAR-10. These hyperparameters are summarized within the respective configuration files. 

\subsection{Datasets Used} \label{sec:datasets}
\begin{table}[h]
\centering
 \begin{tabular}{p{0.12\textwidth}rrrr} 
 \toprule
 \textbf{Dataset} & \textbf{IS}& \textbf{Train}& \textbf{Test} & \textbf{Cat}\\ \midrule
 EMNIST~\cite{cohen_emnist_2017} & $28\times28$ &  112,800 & 18,800 & 47 \\ 
 CIFAR-10~\cite{krizhevsky_cifar} & $32\times32$ & 50,000 & 10,000 & 10 \\ 
 SVHN~\cite{netzer2011reading} & $32\times32$ & 73,257 & 26,032 & 10 \\ 
 \bottomrule
 \end{tabular}
 \caption{Summary of Datasets Used in Experiments. \textbf{IS}: Image Size. \textbf{Train}: Number of Training Images. \textbf{Test}: Number of Testing Images. \textbf{Cat}: Number of Categories}
\label{tab:datasetSummary}
\end{table}

We used three different datasets in our experiments: EMNIST \cite{cohen_emnist_2017}, CIFAR-10 \cite{krizhevsky_cifar}, and SVHN (Street View House Numbers) \cite{netzer2011reading}. We selected these datasets to facilitate baseline comparisons with the original results from the TRUNK model during our reproducibility studies.

Each of these datasets consist of fixed-size images centered around a single object. The EMNIST dataset is a collection of handwritten digits and letters from the English alphabet. It contains 112,800 gray-scaled training images and gray-scaled 18,800 testing images across 47 categories. Each image in the dataset has a dimension of $28\times28$ pixels. The CIFAR-10 dataset consists of 50,000 training images and 10,000 testing images across 10 categories ranging from animals to vehicles. Each image in the dataset has a dimension of $32\times32$ pixels. Finally, the SVHN dataset consists of digital images of house numbers obtained from Google Street View Images. There are 73,257 training images and 26,032 testing images each with a dimension of $32\times32$ pixels across 10 different categories. Table \ref{tab:datasetSummary} gives an overview of the datasets used for our experiments. 

\subsection{Statement of Positionality}
One of the contributors to this paper also served as the principal author of the Tree-Based Unidirectional Neural Network (TRUNK) \cite{goel_modular_2020, TRUNK_goel}. During our research, we collaborated with this author to gain insights and validate findings related to the reproducibility of this case study. For instance, we sought clarification on the source code implementation of TRUNK and discussed the reason for the challenges we encountered in reproducing the results. Through our correspondence with the principal author of TRUNK, we were able to ascertain that our thought process was in the right direction.


\section{Experiments and Results} \label{sec:experiments}
Our experiments involve reproducing our chosen case study. During the analysis, we will enhance the DL software to improve its reproducibility. This section is organized as follows:
\begin{itemize}
    \item \S\ref{sec:softEnv} will explore the replication of the software environment for TRUNK.
    \item \S\ref{sec:preTrainedResults} will assess the validity of the reported results of TRUNK.
    \item \S\ref{sec:trainingReproduceExp} will analyze the reproducibility of training the TRUNK software.
    \item \S\ref{sec:e2eTrain} will explain the significance of providing an end to end training implementation for enhancing the reproducibility of DL software.
    \item \S\ref{sec:groupingVolatilityBatchNorm} exemplifies the need for conducting a sensitivity analysis to facilitate for reproducibility.
    \item \S\ref{sec:trainTransp} emphasizes the significance of disclosing the data processing and training pipelines to reproducibility.
    \item \S\ref{sec:doc} demonstrates the information required to improve the reproducibility of DL software.
\end{itemize}


\subsection{Software Environment Set-Up} \label{sec:softEnv}
The first step in reproducing TRUNK is to replicate their software environment. We will begin this section with analyzing the robustness of the TRUNK software's instructions on replicating their environment and if there is any room for enhancements/improvements to be made. 

The authors of TRUNK provide a manifest \cite{goel_modular_2020} which facilitates the installation of necessary Python libraries using \texttt{pip}. However, an error occurred during the automatic installation of the dependencies listed in the manifest:  ``\textcolor{red}{\texttt{no matching distribution found}}''. We observed that this was due to the manifest's organization; it encompasses not only the essential dependencies but also includes some that the source code does not directly utilize. Some example of these dependencies are \texttt{anaconda-client}, \texttt{blaze}, and \texttt{clyent} to name a few.

Errors with secondary dependencies such as the one we encountered while setting up the environment for TRUNK can occur for a number of reasons, including using different operating systems, conflicts in dependency versions due to installation order, using different computing hardware, or even using different python versions for example. 
To prevent encountering any obstacles similar to this, we recommend only recording the dependencies directly imported into the source code within the manifest. 

Furthermore, the manifest did not specify the CUDA version requisite for integration with PyTorch. This omission would default to the CPU version of PyTorch, when the desired outcome is a version compatible with CUDA drivers. In order to force the CUDA version of PyTorch to be installed in the environment through \texttt{pip}, we provide the link \url{https://download.pytorch.org/whl/cu118} in the manifest to specifically search for and install a version of PyTorch that is compatible with CUDA 12.1. 
Listings \ref{code:requirements.txt}-\ref{code:condaYAML} demonstrates the recommended structure of a manifest to reproduce TRUNK's software environment for \texttt{pip} and \texttt{conda} respectively if the libraries that were directly imported into the TRUNK software were \begin{enumerate*}
    \item[1)] Numpy
    \item[2)] Scipy
    \item[3)] Torch, and
    \item[4)] Torchvision.
\end{enumerate*}

\begin{lstlisting}[
    language=TeX,
    caption=Structure of the requirements.txt manifest to reproduce TRUNK's software environment using \texttt{pip} by only listing the libraries that were directly imported by the developer,
    label={code:requirements.txt},
    basicstyle=\ttfamily\small,
    frame=single,
    xleftmargin=10pt,
    framexleftmargin=5pt,
    breaklines=true
]
--find-links https://download.pytorch.org/whl/cu121
numpy
scipy
torch
torchvision
\end{lstlisting}

\begin{lstlisting}[
    language=TeX,
    caption=Structure of the environment.yaml manifest to reproduce TRUNK's software environment using \texttt{conda} by only listing the libraries that were directly imported by the developer,
    label={code:condaYAML},
    basicstyle=\ttfamily\small,
    frame=single,
    xleftmargin=10pt,
    framexleftmargin=5pt,
    breaklines=true
]
name: trunk
channels:
  - pytorch
  - nvidia
  - defaults
dependencies:
  - numpy
  - python=3.9.18=h955ad1f_0
  - pytorch=2.3.0=py3.9_cuda12.1_cudnn8.9.2_0
  - pytorch-cuda=12.1=ha16c6d3_5
  - torchvision=0.18.0=py39_cu121
  - pip:
    - scipy
\end{lstlisting}

\begin{tcolorbox} [width=\linewidth, colback=blue!20!white, top=1pt, bottom=1pt, left=2pt, right=2pt]
    We \textbf{expand} on the current guidelines regarding the software environment by advocating for a manifest that only lists the primary dependencies with GPU compatibility.
\end{tcolorbox}

\subsection{Inference with Pre-Trained Weights} \label{sec:preTrainedResults}
\begin{table}[h]
\centering
\begin{tabular}{l|c|c|c|}
\cline{2-4}
                                                 & \textbf{EMNIST} & \textbf{CIFAR-10} & \textbf{SVHN} \\ \hline
\multicolumn{1}{|c|}{\textbf{Accuracy {[}\%{]}}} & 85.77           & 91.99             & 96.75         \\ \hline
\end{tabular}
\caption{Conducting inference on the provided pre-trained weights for each dataset}
\label{tab:preTrainedWeights}
\end{table}

Once we have replicated the software environment, the next step is to verify the results reported in the paper. This is done by conducting inference on the provided pre-trained weights. We found that the pre-trained weights correspond to the results reported in the paper \cite{TRUNK_goel}. The availability of these pre-trained weights not only allowed us to assess the validity of this method but also provided a benchmark to compare against when we reproduce the training effort detailed in \S\ref{sec:trainingReproduceExp}. This accessibility to the pre-trained weights enable the verification of reproduction of deep learning models.  

\begin{tcolorbox} [width=\linewidth, colback=orange!20!white, top=1pt, bottom=1pt, left=2pt, right=2pt]
    The current guidelines recommended by various publications suggest the inclusion of pre-trained weights. This guideline is necessary to verify the validity of the novel DL method and to provide a reasonable benchmark to evaluate the reproducibility of training the DL network.
\end{tcolorbox}

\subsection{Reproducing the Training of TRUNK} \label{sec:trainingReproduceExp}
\begin{table}[h]
\centering
\begin{tabular}{c|c|c|c|}
\cline{2-4}
\textbf{}                                                & \textbf{EMNIST} & \textbf{CIFAR-10} & \textbf{SVHN} \\ \hline
\multicolumn{1}{|c|}{\textbf{Train Results {[}\%{]}}}      & 63.62           & --.--             & 98.22         \\ \hline
\multicolumn{1}{|c|}{\textbf{Org. Results {[}\%{]}}} & 85.77           & 91.99             & 96.75         \\ \hline
\end{tabular}
\caption{Comparing the results obtained after reproducing the training efforts \textbf{(train results)} with the original \textbf{(org.)} results obtained from the pre-trained weights}
\label{tab:trainResultsvsPreTrainedWeights}
\end{table}
Now that we have confirmed the validity of the results obtained and the methodology proposed by the original authors of TRUNK, the next step is to assess whether or not training the entire deep learning network will yield consistent results. Using the training scripts provided by the authors of TRUNK \cite{goel_modular_2020}, we attempt to reproduce the training effort and evaluate the similarity of validation accuracy results to those achieved with the provided pre-trained weights. The accuracies of the pre-trained weights and the accuracies achieved after training TRUNK from scratch are shown in Table \ref{tab:trainResultsvsPreTrainedWeights}. 

While we achieved comparable accuracies with the TRUNK software on the SVHN dataset after training from scratch, we did encounter several challenges:
\begin{enumerate}
\item The accuracy for the EMNIST dataset could not be replicated when training the network from scratch.
\item For the CIFAR-10 dataset, critical details were missing, including hyperparameters and the data-processing pipeline. Which is why we were unable to train TRUNK on CIFAR-10. 
\item Documentation for effectively training TRUNK was limited, resulting in some confusion regarding the execution of the source code.
\end{enumerate}

In addition to the challenges listed above, we also noticed that a pre-defined tree structure was provided to us for both the EMNIST and SVHN dataset. Using the source code provided, we were able to only train each node within this pre-defined tree. However, a distinctive feature of the TRUNK network is the construction of the tree structure itself, alongside training each node. Unfortunately, this building aspect was not represented in the source code, preventing us from reproducing the provided tree structure as we trained the network on the given dataset. As a result, we were unable to achieve end-to-end reproduction of the training of the TRUNK network.

The subsequent sections will detail our approaches to addressing the challenges outlined above and enhancing the reproducibility of the TRUNK network through improvements to the TRUNK software.

\begin{figure}[h]
\begin{center}
\centering
\subfigure[]{\includegraphics[width=0.5\textwidth, height=3cm]{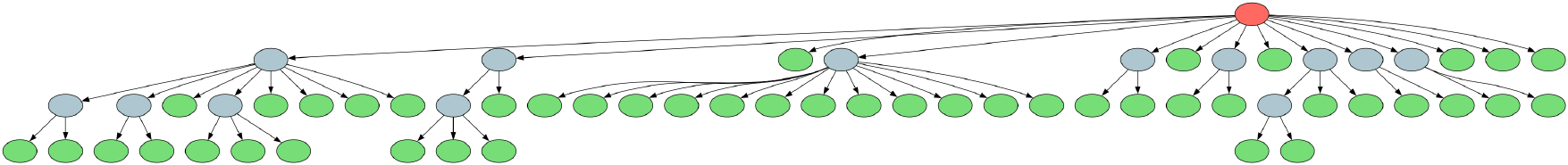}
}
\subfigure[]{\includegraphics[width=0.5\textwidth, height=3cm]{Figures/TRUNK/colorTrees/emnist_colored_tree.pdf}
}
\end{center}
\caption{Comparison of the EMNIST (a) Tree Structure Developed by TRUNK Authors with (b) Our Reproduced Tree Structure. The red node is the root node of the tree, the gray nodes are the supergroups, and the green nodes are the leaf nodes.}
\label{fig:ourEMNISTvsAuthorEmnist}
\end{figure}

\subsection{End to End Training Implementation} \label{sec:e2eTrain}
\begin{figure*}[h]
\centering
\includegraphics[width=0.70\textwidth, height=0.39\textwidth]{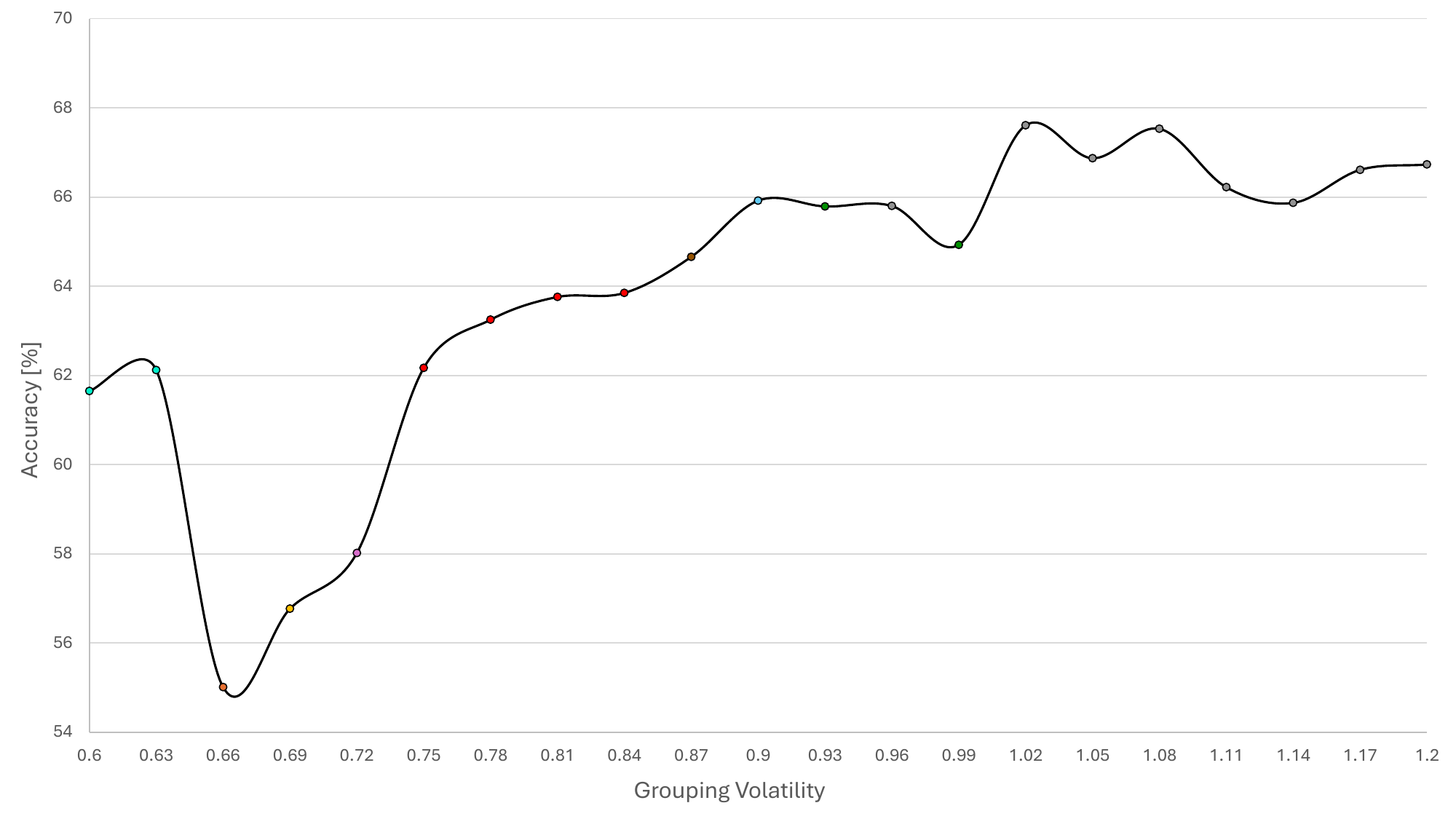}
\caption{Ablation study analyzing the sensitivity of the grouping volatility hyperparameter for CIFAR-10. Each data point color represents a specific tree built.} 
\label{fig:ablationBatch}
\end{figure*}

\begin{figure*}[h]
\centering
\includegraphics[width=0.70\textwidth, height=0.39\textwidth]{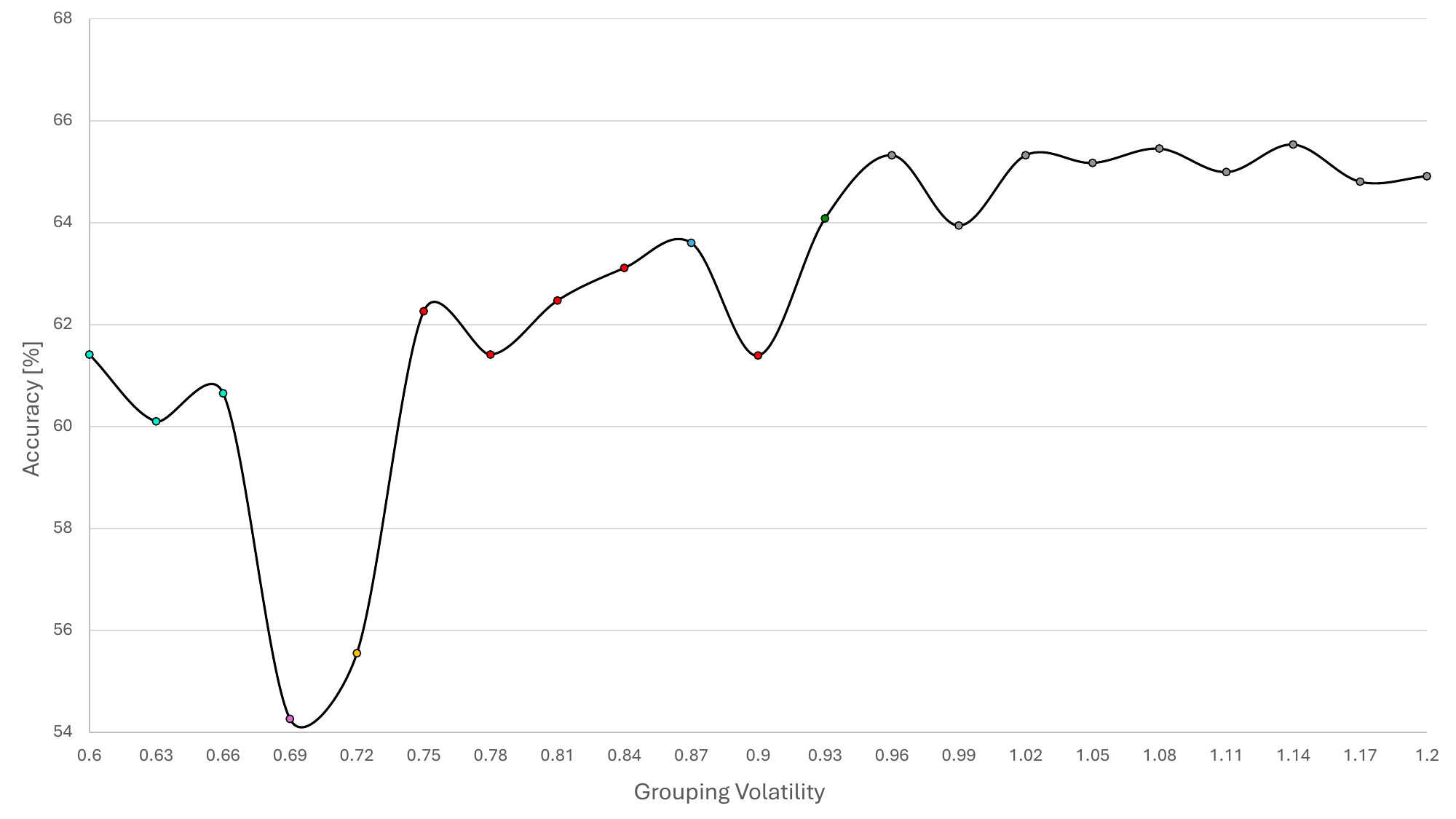}
\caption{Analyzing the sensitivity of the grouping volatility parameter by changing the original implementation to include layer normalization. Each data point color represents a specific tree built.} 
\label{fig:layerNormAblation}
\end{figure*}

End-to-end training ensures that every step of the learning process—from initial data processing to final output generation—is included within a single, unified training pipeline. This completeness is essential for reproducibility because it allows for the comprehensive testing of all factors and not just the final performance of the architecture.  

\begin{tcolorbox} [width=\linewidth, colback=orange!20!white, top=1pt, bottom=1pt, left=2pt, right=2pt]
    The current guidelines suggest that the source code should reproduce the theoretical framework presented in the paper. This is crucial for accurately reproducing and evaluating the characteristics of the new deep learning method, as highlighted by our challenges in verifying the tree structure of TRUNK.
\end{tcolorbox}

In monolithic networks such as ResNet \cite{he_deep_2015}, which is originally trained on the ImageNet \cite{krizhevsky_imagenet_2012} dataset featuring 1,000 categories, modifications are minimal when adapting to different datasets. For instance, to train ResNet on the CIFAR-10 \cite{krizhevsky_cifar} dataset, which contains only 10 categories, we simply adjust the output features layer to match the 10 categories while the rest of the layers remain unchanged. Typically, changes in monolithic networks are confined to the input or output layers based on the dataset requirements, yet the core architecture remains consistent. This is why the end-to-end training of monolithic networks primarily involves training a single extensive network and assessing its overall performance.

\begin{figure*}[h]
\begin{center}
\centering
\subfigure[]{\includegraphics[width=0.65\textwidth]{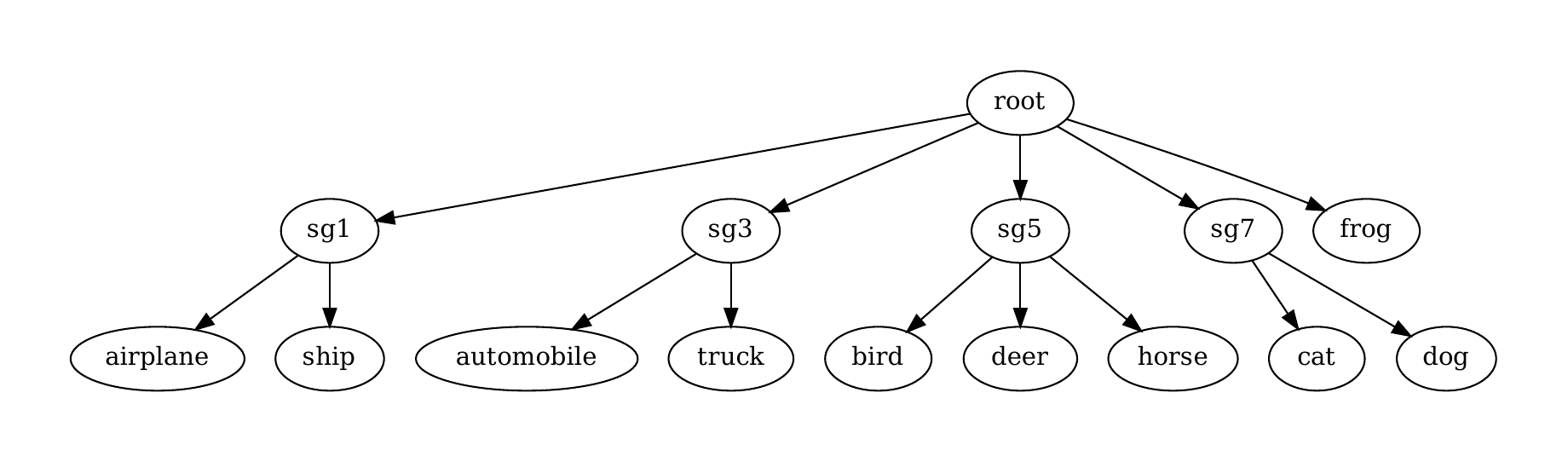}
}
\subfigure[]{\includegraphics[width=0.65\textwidth]{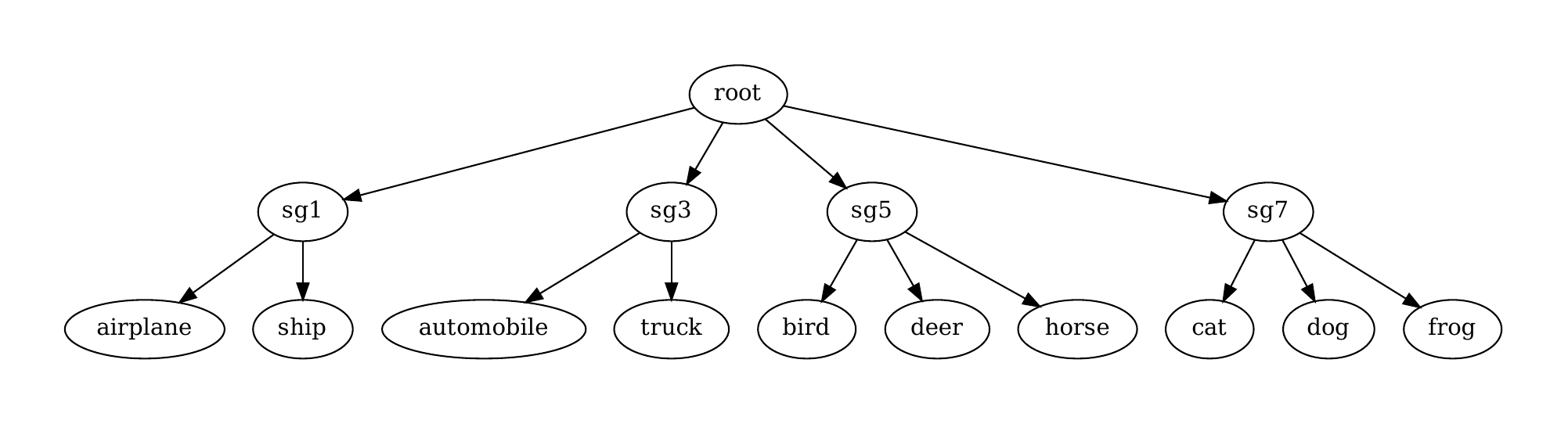}
}
\end{center}
\caption{Analyzing how using different NVIDIA A100 GPUs and the same hyperparameters can change the structure and performance of a model, with (a) Structure of the Tree from the First Execution of the Re-implemented TRUNK software with an accuracy of 81.53\% and (b) the Reproduced Results with an accuracy of 79.72\%. }
\label{fig:reproResults}
\end{figure*}

For unique and complex architectures like TRUNK, which adapt their structure based on the dataset due to the visual similarity criteria, as illustrated in Figure \ref{fig:treeDataset}, the situation is different. If this adaptive characteristic is not embedded in the source code, it can lead to irreproducible results, as full end-to-end training is not achieved. In such cases, the source code may enable the reproduction of network performance, but it fails to allow for the reproduction of the network’s structural design, which is a crucial aspect of TRUNK. Despite having a pre-defined tree structure available, we faced difficulties in determining whether we could reproduce the same tree structure.

To enhance the reproducibility of TRUNK, we have integrated this adaptive structure-building feature directly into the training code, moving away from the use of a pre-defined tree structure and a fixed set of nodes. This integration allows us to not only enable the reproduction of the performance of TRUNK but also reproduce the structure of the tree as well. 

\begin{table}[h]
\centering
\begin{tabular}{c|c|c|c|}
\cline{2-4}
\textbf{}  & \textbf{EMNIST} & \textbf{CIFAR-10} & \textbf{SVHN} \\ \hline
\multicolumn{1}{|c|}{\textbf{Our Results {[}\%{]}}}                 & 84.30           & 67.61             & 90.24         \\ \hline
\multicolumn{1}{|c|}{\textbf{Org. Results {[}\%{]}}} & 85.77           & 91.99             & 96.75         \\ \hline
\end{tabular}
\caption{Comparing the results obtained after reproducing the build and training effort of TRUNK \textbf{(our results)} with the original \textbf{(org.)} results obtained from the pre-trained weights}
\label{tab:buildResultsvsPreTrainedWeights}
\end{table}

Table \ref{tab:buildResultsvsPreTrainedWeights} compares the results obtained from the pre-trained weights with those achieved after reproducing the TRUNK tree structure and training each individual node of the tree. After integrating the tree build characteristic of TRUNK, we were able to improve the performance of TRUNK on the EMNIST dataset by 20.68\%. But the tree structure obtained for the EMNIST dataset differed from the original structure reported by the authors as demonstrated in Figure \ref{fig:ourEMNISTvsAuthorEmnist}. We will explore more on the sensitivity of the tree structures of TRUNK in \S\ref{sec:groupingVolatilityBatchNorm} and how this impacts the reproducibility of deep learning software. 

For the CIFAR-10 dataset, since crucial details of the hyperparameters were missing, we make an assumption to use the same hyperparameters and the data-processing pipeline from training the SVHN dataset. Despite integrating the build characteristic and reproducing the training of TRUNK end-to-end, there was still nearly a 25\% difference in accuracy and a change in the tree design architecture from the pre-trained weights provided. This disparity is the result of not having crucial details such as the training regime and the data processing pipeline used for training TRUNK on the CIFAR-10 dataset. \S\ref{sec:trainTransp} will explore the necessity of outlining the training regime and the data-processing pipeline to improve the reproducibility of deep learning software. 

\subsection{Sensitivity Analysis of Hyperparameters} \label{sec:groupingVolatilityBatchNorm}
\begin{figure}[h]
\begin{center}
\centering
\subfigure[]{\includegraphics[width=0.5\textwidth]{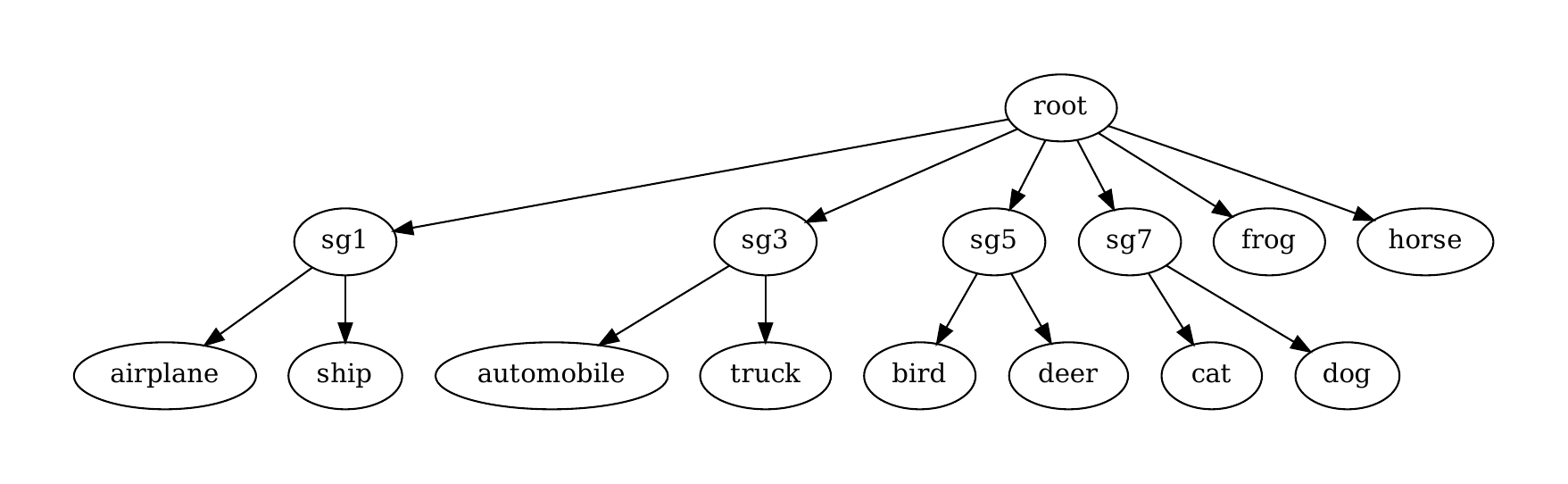}
}
\subfigure[]{\includegraphics[width=0.5\textwidth]{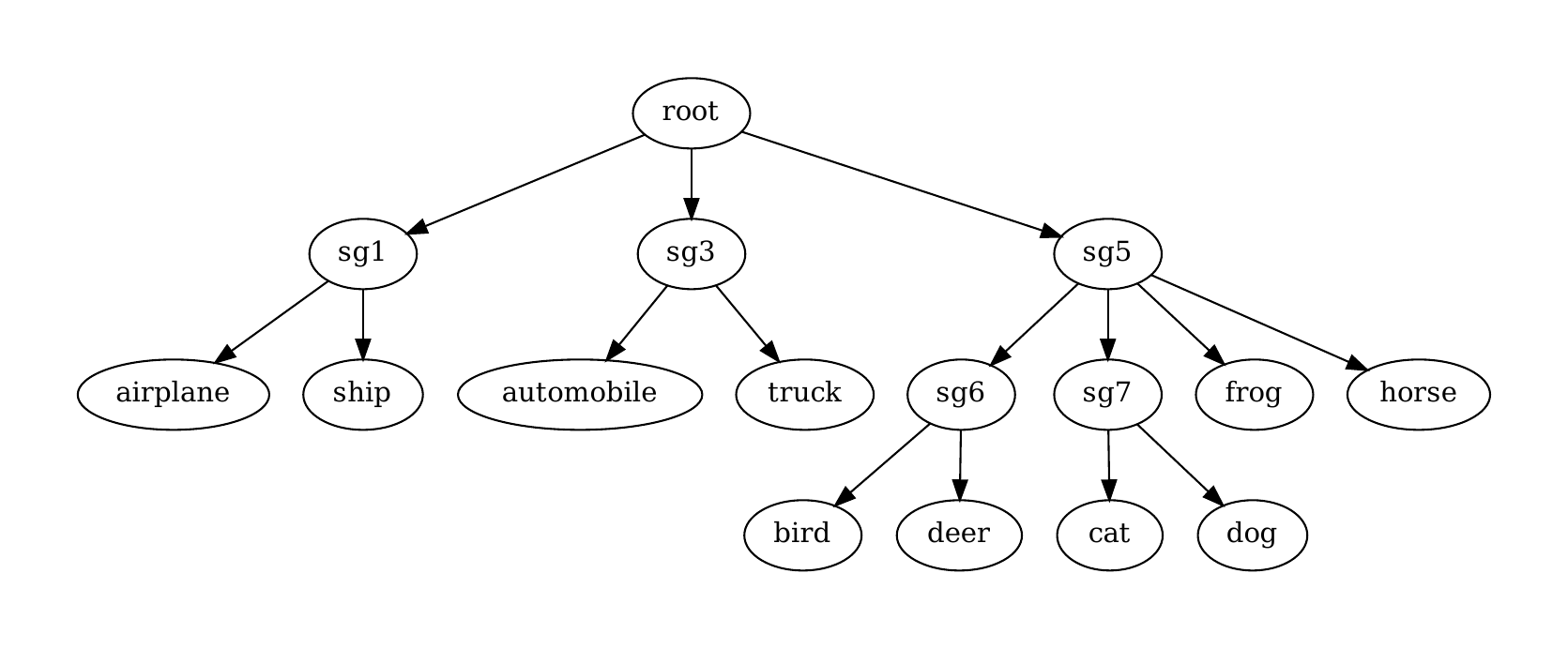}
}
\end{center}
\caption{Differences in the TRUNK tree structures between (a) the structure achieving the highest accuracy and (b) the reported tree structure by the original authors of TRUNK.}
\label{fig:HighestvsReported}
\end{figure}

To understand why there is a disparity between the tree configurations achieved using the pre-trained weights and the tree configuration achieved after reproducing the training of TRUNK, we will conduct an ablation study to study the sensitivity of the \textit{grouping volatility} hyperparameter on the CIFAR-10 dataset. The \textit{grouping volatility} is a unique hyperparameter to TRUNK and is the threshold that determines whether two categories should be clustered \cite{goel_modular_2020, TRUNK_goel}. This hyperparameter influences the design of the tree structure for TRUNK. We keep every other hyperparameter constant in our sensitivity analysis. These hyperparameters were selected based on the parameters used by the original authors for the SVHN and EMNIST datasets.   

\begin{figure}[h]
\begin{center}
\centering
\subfigure[]{\includegraphics[width=0.5\textwidth]{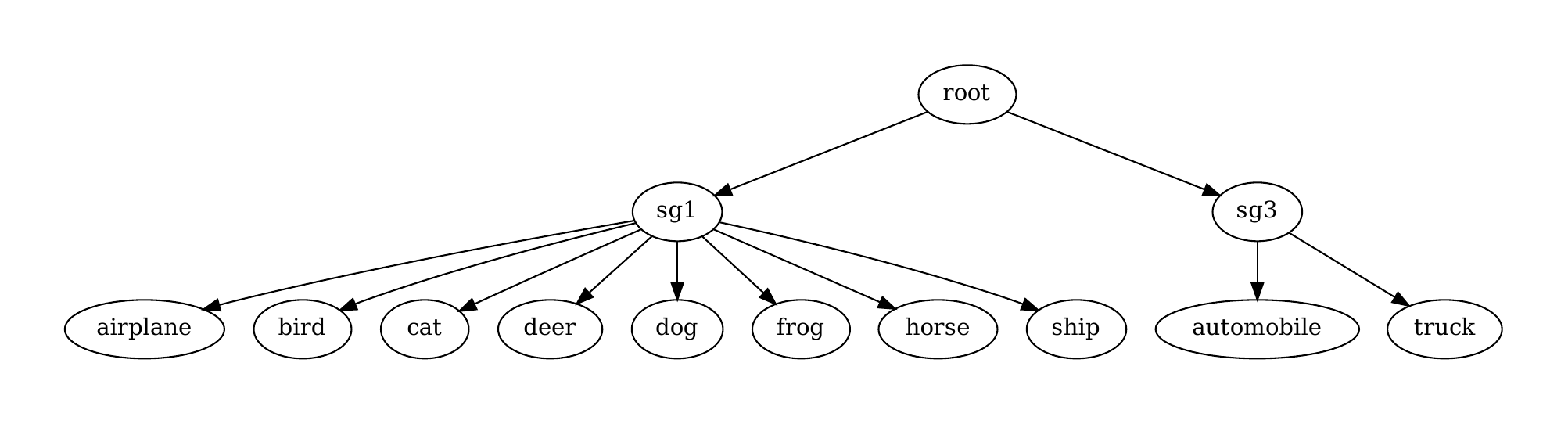}
}
\subfigure[]{\includegraphics[width=0.50\textwidth]{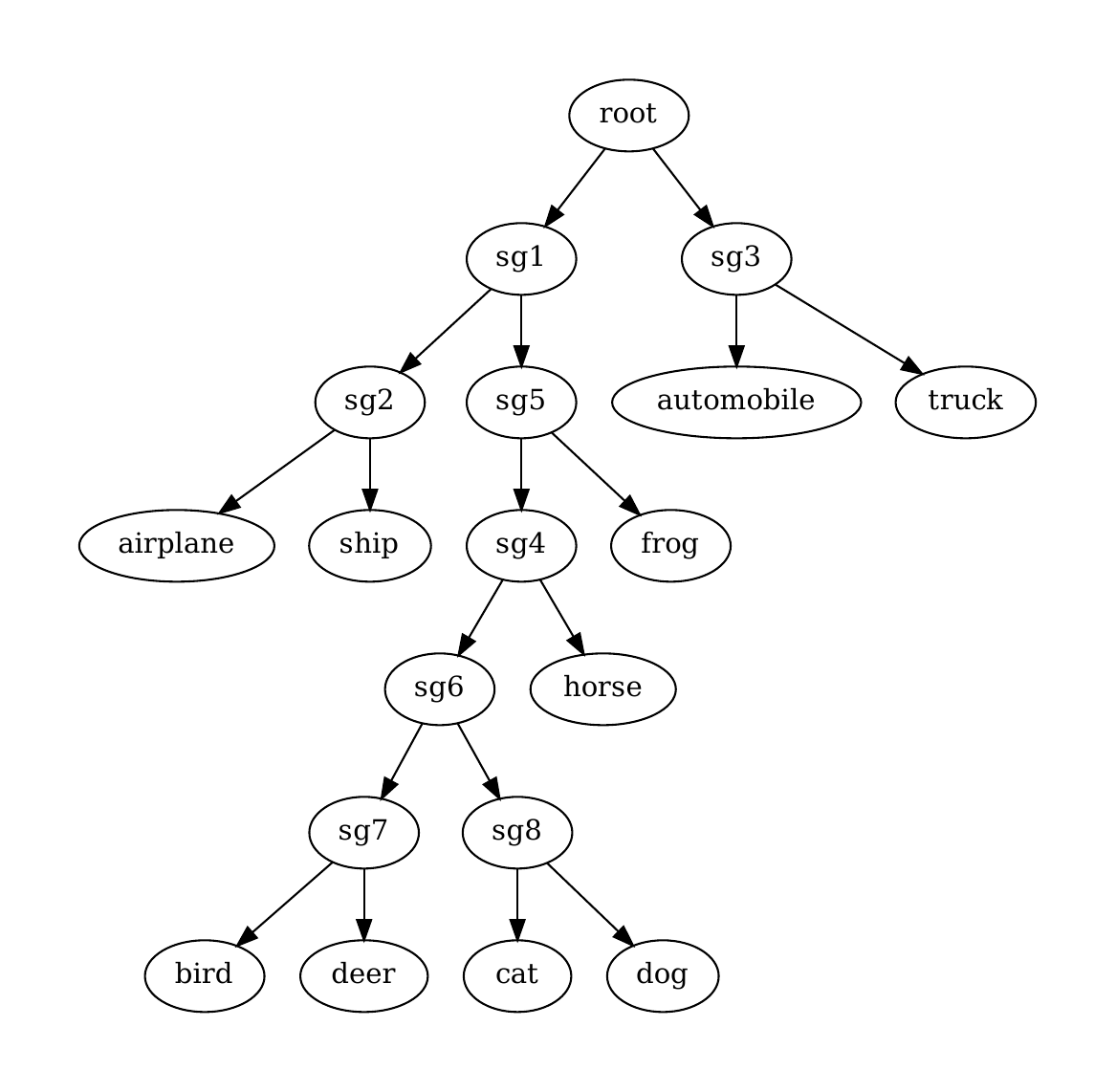}
}
\end{center}
\caption{A slight change in a hyperparameter—such as adjusting the grouping volatility from (a) 0.63 to (b) 0.66—can drastically affect how the network is constructed and performs, as demonstrated in our case study. The deeper tree underperformed compared to the shallower one, suggesting that an optimal balance between the depth and width of the tree is necessary.}
\label{fig:accTrees}
\end{figure}

We increment the grouping volatility hyperparameter by 0.03 from 0.60 to 1.20 and record the overall accuracy as illustrated in Figure \ref{fig:ablationBatch}. Each data point color represents a specific tree structure, with the red representing the tree structure the original authors \cite{goel_modular_2020, TRUNK_goel} achieved. The original authors of TRUNK set $\text{grouping volatility} = 1$, but in our attempt to reproduce their results, we observed a significantly different tree structure at the same grouping volatility, as depicted in Figure \ref{fig:ablationBatch} and Figure \ref{fig:HighestvsReported}.

\begin{figure*}[h]
\centering
\includegraphics[width=0.70\textwidth]{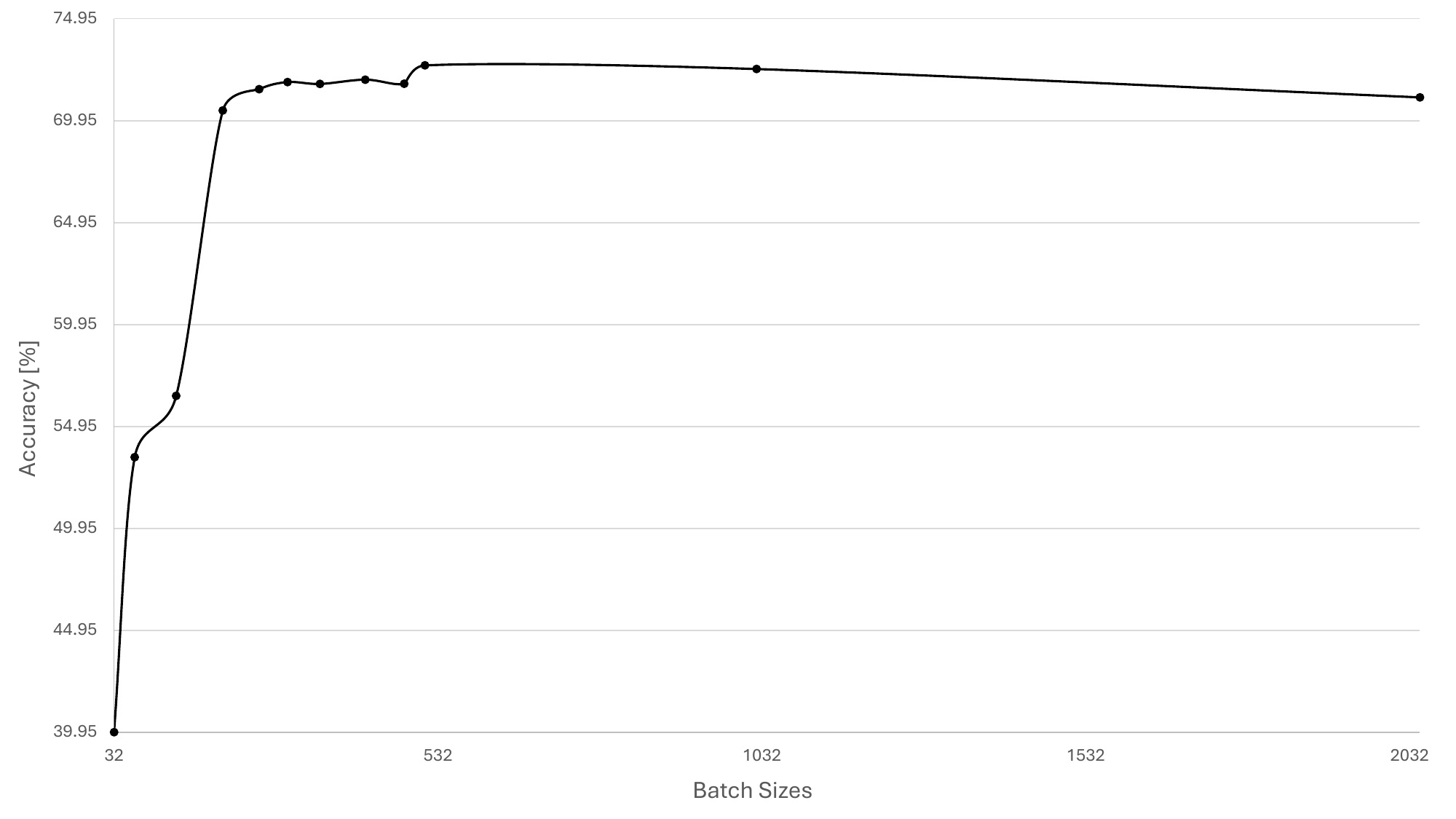}
\caption{Ablation over the batch sizes with revised data augmentations \cite{cifar10_pytorch_github} and the same tree network as the authors for the CIFAR-10 dataset.} 
\label{fig:batchAblate}
\end{figure*}

Using a grouping volatility of 1.02, we achieved the highest accuracy with the TRUNK model on the CIFAR-10 dataset, reaching 67.61\%. However, reducing the grouping volatility by 0.18 from this optimal setting resulted in a decrease in accuracy to 63.85\% with the tree structure that was reported by the original authors, as illustrated in Figure \ref{fig:HighestvsReported}.

From Figure \ref{fig:ablationBatch}, we also see that there was a drop in accuracy by 7\% over a 0.03 difference in grouping volatility from 0.63 to 0.66. The structure of the tree between these points are also very different with one being deeper than the other (Figure \ref{fig:accTrees}). This increase in depth contributed to this drop in accuracy.  

The initial hypothesis as to why this sensitivity in the grouping volatility may exist was due to the stochastic nature of using batch normalization. That is why we modified the architecture to replace all batch normalization layers with layer normalization \cite{layerNorm}. 
As depicted in Figure \ref{fig:layerNormAblation}, even after using layer normalizations, the general trend of the grouping volatility sensitivity remained.

The analysis suggests that grouping volatility is independently a sensitive hyperparameter. It shows a clear trend: increasing the grouping volatility results in more groupings per node, whereas decreasing it leads to fewer groupings. This adjustment influences whether the tree structure becomes deeper or wider.

Our ablation study reveals that establishing a strong foundation for reproducibility is not merely about selecting optimal hyperparameters; it also requires a comprehensive understanding of the model’s behavior across a wide range of scenarios. Through our ablation experiments, we observed that increasing the grouping volatility leads to a higher number of groupings per node, whereas decreasing the grouping volatility results in fewer groupings per node.

Understanding the sensitivity of the grouping volatility hyperparameter is useful due to the inherent variability in deep learning research caused by pseudo-random number generation. For instance, as shown in Figure \ref{fig:reproResults}(b), the tree structure differs from that in Figure \ref{fig:reproResults}(a) and achieves lower accuracy, even though the same seed, training regime, and data processing pipeline were used from for Training Regime 6 outlined in Table \ref{tab:summaryDataAug}. This difference in results exists due to how GPUs handle floating-point calculations differently \cite{gemmNVIDIA} or due to the influence of the randomness in the software \cite{semmelrock2023reproducibility, seed} which was made prominent due to the sensitive hyperparameters involved with training TRUNK. Based on our ablation study findings in Figure \ref{fig:ablationBatch}, we know that increasing the grouping volatility parameter is necessary to reproduce the tree structure and accuracy observed in Figure \ref{fig:reproResults}(a).

\begin{tcolorbox} [width=\linewidth, colback=blue!20!white, top=1pt, bottom=1pt, left=2pt, right=2pt]
    We \textbf{expand} the current guidelines by advocating for sensitivity analysis to comprehend the model's behavior across a wide spectrum of parameters. This approach enables fellow researchers to make the necessary adjustments to closely reproduce the original implementation by the authors, despite the inherent non-determinism and randomness.
\end{tcolorbox}



\subsection{Transparency of the Training and Data-Processing Pipeline} \label{sec:trainTransp}
\begin{figure}[h]
\begin{center}
\centering
\subfigure[]{\includegraphics[width=0.20\textwidth]{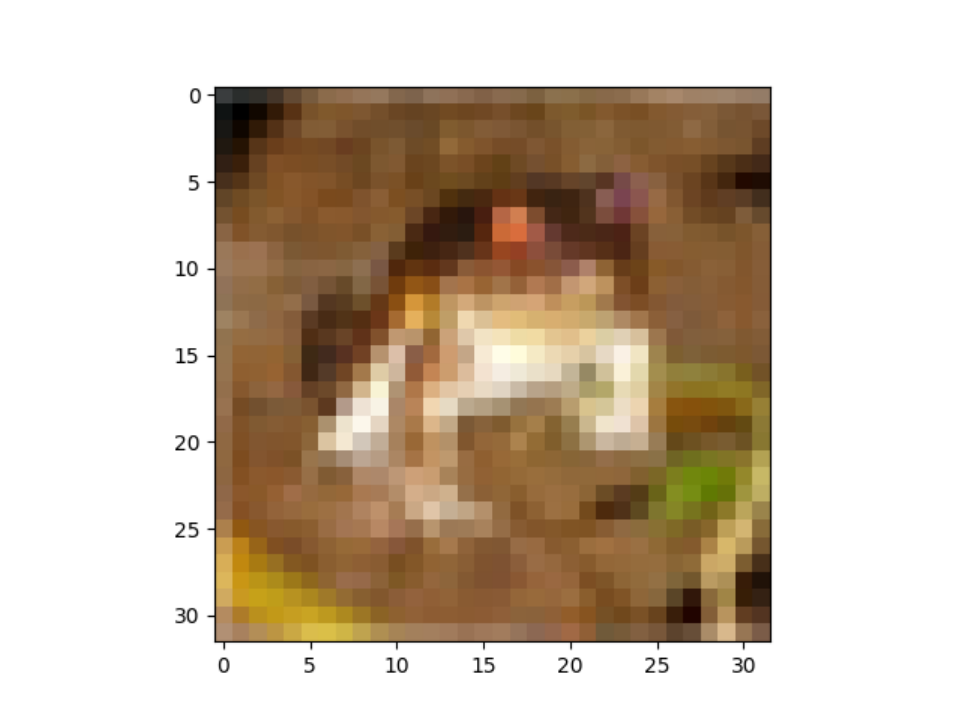}
}
\subfigure[]{\includegraphics[width=0.20\textwidth]{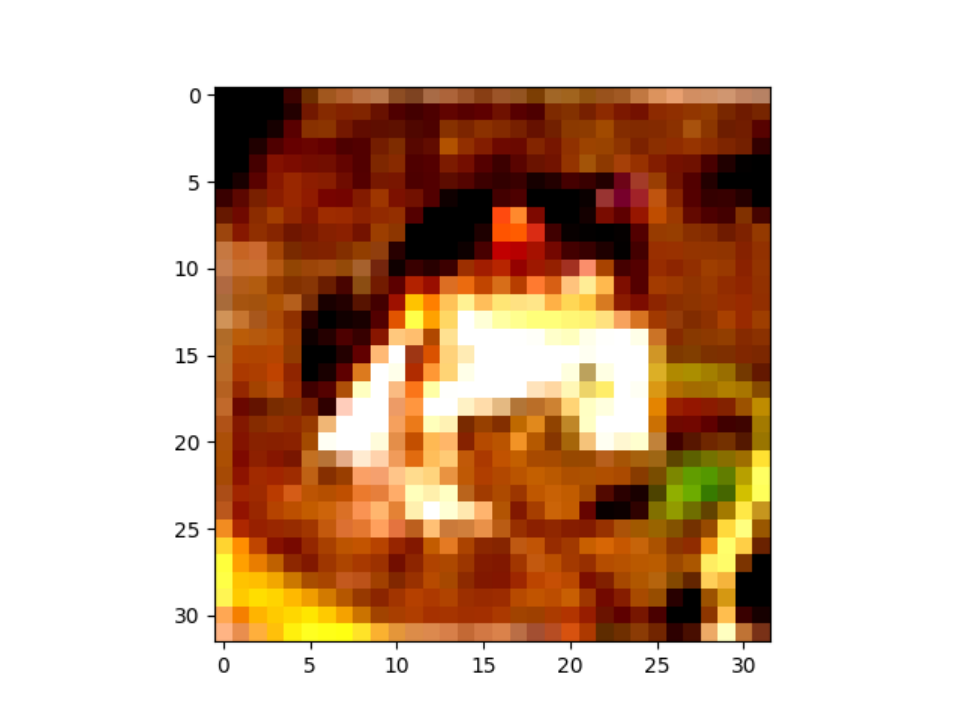}
}
\subfigure[]{\includegraphics[width=0.20\textwidth]{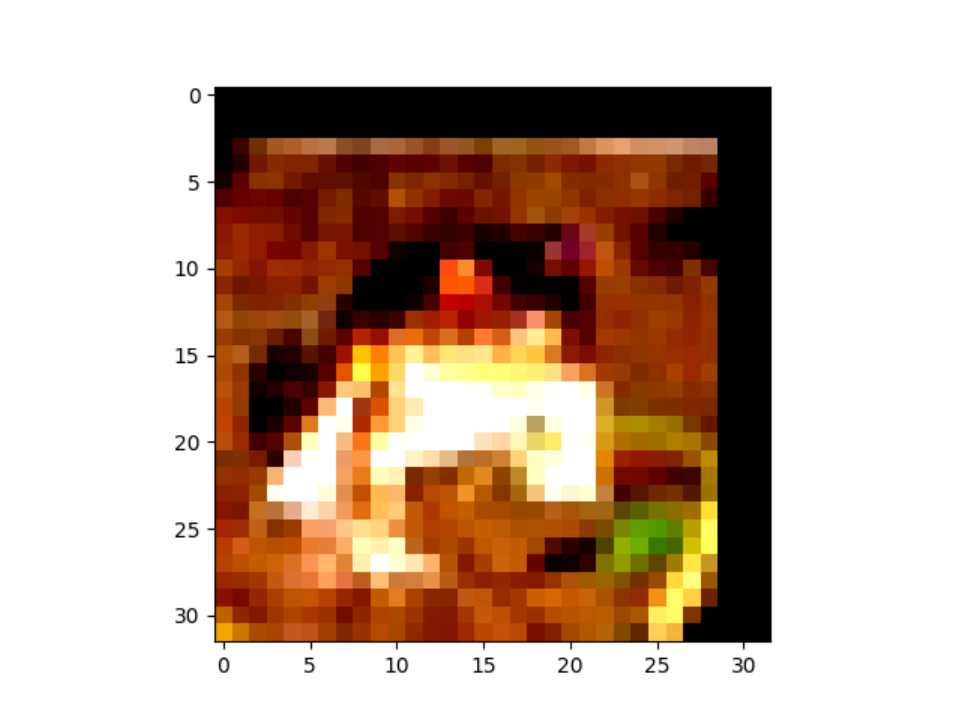}
}
\subfigure[]{\includegraphics[width=0.20\textwidth]{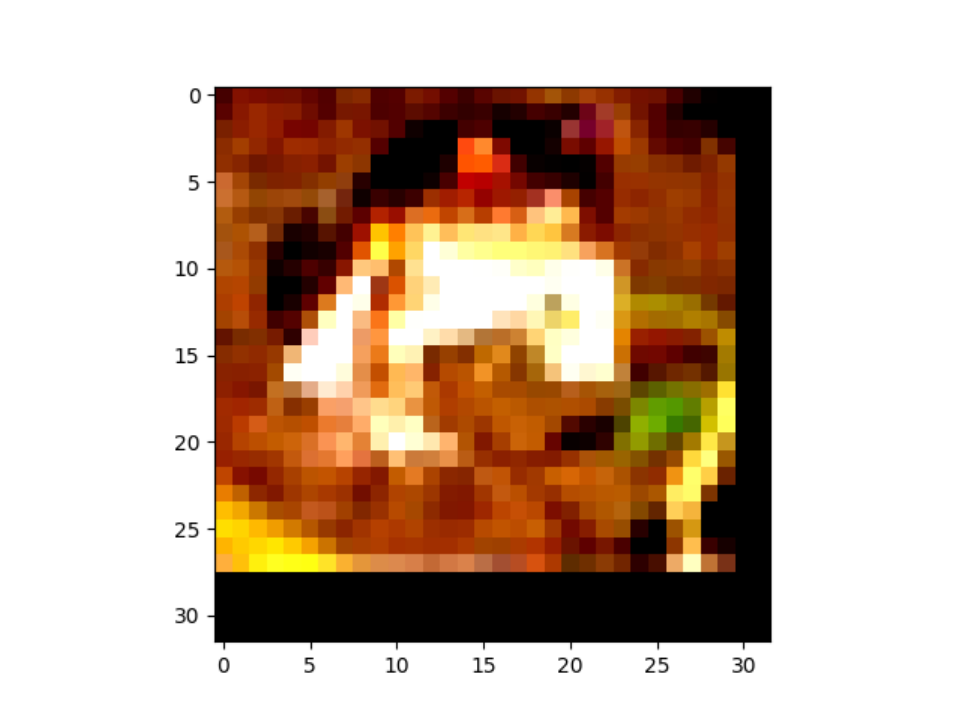}
}
\end{center}
\caption{Example of different augmentation techniques to illustrate the transformations undergone by the data sample: [Left to Right, Top to Bottom] (a) The original image before augmentations (b) The image after normalization (c) The image after applying random crop (d) The image after applying random horizontal flip.}
\label{fig:transformations}
\end{figure}

In this section we will look into the necessity of being transparent with the training and data-processing pipeline. This section is organized as follows:
\begin{itemize}
    \item \S\ref{sec:sigTrain} outlines the significance of disclosing the training regimen
    \item \S\ref{sec:sigData} describes the methods used to enhance the training and data-processing pipelines for the TRUNK network.
\end{itemize}
 
\subsubsection{Significance of Disclosing the Training Recipe} \label{sec:sigTrain}
Studies have shown that the training recipe and the data-processing pipeline applied to a dataset, significantly influences the performance of the model as it helps in preventing over-fitting and allows the model to generalize better on unseen data \cite{deit3, vitDataAug}. As illustrated in Figure \ref{fig:transformations}, the final augmented image is noticeably distinct from the original image. This divergence can result in varied performance outcomes across different architectures due to the final transformed image possessing features that may be absent in the original image and useful for model learning. This is demonstrated by our results in Table \ref{tab:summaryDataAug} for the CIFAR-10 dataset using the TRUNK network. 

Another parameter that influences the performance of a model is known as the batch size and batch normalization \cite{batchNorm}. 
In Figure \ref{fig:batchAblate}, we observe that batch size significantly affects the model’s performance, with the architecture under-performing at smaller batch sizes. Gradually, as the batch size increases, we identify an optimal value that yields the highest accuracy, making it an important factor to disclose.  

In addition to providing the algorithm to train a novel deep learning, we need to also disclose the architecture design, and training regime (TR) used \cite{natureTransp} to achieve similar results as demonstrated by the results in Table \ref{tab:summaryDataAug}.

We modify the training and data-processing pipeline used from the initial assumption made to several other recommended practices for the CIFAR-10 dataset. These changes are outlined in Table \ref{tab:summaryDataAug}. By revising the hyperparameters, we see nearly a 14\% jump in accuracy from TR1 to TR6. By conducting this experiment, we highlight the significance of documenting the training regime and data-processing pipeline used to train the deep learning network \cite{natureTransp, semmelrock2023reproducibility, torchDistill}. 

\begin{table*}[h]
\centering
\begin{tabular}{clllccccccclll}
\toprule
\textbf{TR}       & \textbf{LR} & \textbf{BS} & \textbf{GV} & \textbf{RC} & \textbf{RHF} & \textbf{RR} & \textbf{CJ} & \textbf{RA.} & \textbf{CO} & \textbf{TT} & \textbf{Norm. Mean} & \textbf{Norm. Stdev} & \textbf{Acc.}\\ \midrule
\textbf{1} & $1\mathrm{e}{-4}$ & 1024 & 1.02 & $\times$ & $\times$ & $\times$ & $\times$ & $\times$ & $\times$  &  \checkmark & {[}0.50, 0.50, 0.50{]} & {[}0.50, 0.50, 0.50{]} & 67.61\\ 
\textbf{2}  & $1\mathrm{e}{-3}$ & 512 & 0.79 & \checkmark & \checkmark & $\times$ & $\times$ & $\times$ & $\times$ & \checkmark & {[}0.49, 0.48, 0.45{]} & {[}0.20, 0.20, 0.20{]} & 72.66\\ 
\textbf{3}  & $2.36\mathrm{e}{-4}$ & 512 & 0.93 & \checkmark & \checkmark & $\times$ & $\times$ & $\times$ & \checkmark & \checkmark & {[}0.49, 0.48, 0.45{]} & {[}0.25, 0.24, 0.26{]} & 76.53\\ 
\textbf{4}  & $2.36\mathrm{e}{-4}$ & 512 & 0.78 & $\times$ & \checkmark & \checkmark & \checkmark & $\times$ & $\times$ & \checkmark & {[}0.49, 0.48, 0.45{]} & {[}0.25, 0.24, 0.26{]} & 78.18\\ 
\textbf{5}  & $2.36\mathrm{e}{-4}$ & 512 & 0.88 & $\times$ & $\times$ & $\times$ & $\times$ & \checkmark & $\times$ & \checkmark & {[}0.49, 0.48, 0.45{]} & {[}0.25, 0.24, 0.26{]} & 79.15\\ 
\textbf{6}  & $2.36\mathrm{e}{-4}$ & 500 & 0.88 & $\times$ & $\times$ & $\times$ & $\times$ & \checkmark & \checkmark & \checkmark & {[}0.49, 0.48, 0.45{]} & {[}0.25, 0.24, 0.26{]} & 81.53\\ 
\bottomrule
\end{tabular}
\caption{Comparison of the initial and revised training/data-processing pipelines for the TRUNK architecture on the CIFAR-10 dataset. The transition from the original pipeline in the first recipe (inspired by TRUNK’s hyperparameters for SVHN) to the sixth recipe yields an 13.92\% boost in accuracy, underscoring the pivotal role of detailed reporting for reproducibility. Abbreviations: \textbf{TR}: Training Regime. \textbf{LR}: Learning Rate. \textbf{BS}: Batch Size. \textbf{GV}: Grouping Volatility. \textbf{RC}: Random Crop. \textbf{RHF}: Random Horizontal Flip. \textbf{RR}: Random Rotation. \textbf{CJ}: Color Jitter. \textbf{RA.}: Random Augmentation \cite{randAug}. \textbf{CO}: CutOut \cite{cutout}. \textbf{TT}: ToTensor. \textbf{Norm. Mean}: Normalized Mean. \textbf{Norm. Stdev}: Normalized Standard Deviation. \textbf{Acc.}: Accuracy.}
\label{tab:summaryDataAug}
\end{table*}

\begin{tcolorbox} [width=\linewidth, colback=orange!20!white, top=1pt, bottom=1pt, left=2pt, right=2pt]
    As current guidelines recommend, it is important to document the data processing and training pipelines, along with the sensitivity analysis conducted. This approach enables an understanding of the parameters originally selected by the authors of the novel deep learning method as a starting point before conducting the sensitivity analysis, facilitating for reproducibility.
\end{tcolorbox}

\subsubsection{Enhancing the Transparency of the Data-Processing and Training Pipeline} \label{sec:sigData}
The current TRUNK software \cite{TRUNK_goel} discloses their training and data processing stage scattered deep within their source code. To improve the clarity of this step and provide high-level summaries of experiments, we have introduced a PyYAML configuration file. This file meticulously outlines all parameters involved in the training and data processing pipeline, drawing inspiration from Matsubara's TorchDistill approach \cite{torchDistill}. Listings \ref{config:dataset}-\ref{config:training} is a sample of what these configuration files look like. 

The configuration file in Listing \ref{config:dataset} provides information from the batch size used to the transformations applied to each dataset. We also list the seed value in the configuration file to limit the randomness in the software in Listing \ref{config:dataset}. 

\begin{lstlisting}[
    language=TeX,
    caption=Enhancing the clarity and transparency of TRUNK's data processing pipeline for the SVHN dataset using a PyYAML configuration file,
    label={config:dataset},
    basicstyle=\ttfamily\small,
    frame=single,
    xleftmargin=10pt,
    framexleftmargin=5pt,
    breaklines=true
]
seed: 42
dataset:
train:
params:
  batch_size: 16
  num_workers: 2
  shuffle: True
transform:
- type: ToTensor
- type: Normalize
  params:
    mean:
    - 0.500
    - 0.500
    - 0.500
    std:
    - 0.500
    - 0.500
    - 0.500
validation:
params:
  batch_size: 16
  num_workers: 2
  shuffle: True
transform:
- type: ToTensor
- type: Normalize
  params:
    mean:
    - 0.500
    - 0.500
    - 0.500
    std:
    - 0.500
    - 0.500
    - 0.500
test:
params:
  batch_size: 1
  num_workers: 2
  shuffle: True
\end{lstlisting}

\begin{lstlisting}[
    language=TeX,
    caption=Enhancing the clarity and transparency of TRUNK's training pipeline for the SVHN dataset using a PyYAML configuration file,
    label={config:training},
    basicstyle=\ttfamily\small,
    frame=single,
    xleftmargin=10pt,
    framexleftmargin=5pt,
    breaklines=true
]
loss: 
- type: NLLLoss
grouping_volatility: 0.70
lr_scheduler: 
- type: CosineAnnealingLR
params:
T_max: 10
eta_min: 0
optimizer: 
- type: Adam
params:
lr: 0.005
weight_decay: 0.0005
epochs: 20
\end{lstlisting}

Similarily, the configuration file for the training pipeline in Listing \ref{config:training} details the specific regime and hyperparameters (i.e. the type of optimizer, learning rate scheduler, learning rate, weight decay, loss function) used to train the network.
By following this standard, we not only provide an overview of the data processing and training pipelines, but we also allow for modifications to be made to conduct experiments without changing the underlying source code.

\subsection{Quality Documentation} \label{sec:doc}
Reproducibility for machine learning research should extend beyond the concept of reproducing the results when using the same data and analytical tools. It should also be about making the research accessible and understandable to fellow researchers. Proper documentation is necessary to be able to reproduce the reported results \cite{ml_chem, pineau2020, Gundersen_Kjensmo_2018, johnson2020datascience}. An example of documentation provided alongside the source code is a README markdown file. According to Stojnic \cite{johnson2020datascience}, the README file should be structured as follows to foster an intuitive transfer of information:

\begin{enumerate}
    \item[(1)] Provide a summary of the research paper.
    
    \item[(2)] Instructions on how to replicate the software environment used by the authors by providing a sample command line argument as shown in Listing \ref{code:installDependencies} using a specific package management tool like \texttt{pip} \cite{pip} or \texttt{conda} \cite{conda}. \newline

\begin{lstlisting}[
    language=TeX,
    caption=Example of providing instructions to install the software dependencies using \texttt{pip} or \texttt{conda},
    label={code:installDependencies},
    basicstyle=\ttfamily\small,
    frame=single,
    xleftmargin=10pt,
    framexleftmargin=5pt,
    breaklines=true
]
# To install software dependencies using Pip
pip install -r requirements.txt

# To install software dependencies using Conda
conda env create -f environment.yml
conda activate mnn
\end{lstlisting}
\end{enumerate}

\begin{enumerate}
    \item[(3)] Instructions on how to train the network by providing the command line argument used to execute the training script. The example shown in Listing \ref{code:trainInstructions} explains the command used to execute the training script for TRUNK. It demonstrates the additional arguments required to train the network such as the type of dataset being used, and what deep neural network design (MobileNet \cite{howard_mobilenets_2017} or VGG \cite{simonyan_very_2014}) we are using.

    \begin{lstlisting}[
    language=TeX,
    caption=Example of providing instructions on how to execute the training for the TRUNK network,
    label={code:trainInstructions},
    basicstyle=\ttfamily\small,
    frame=single,
    xleftmargin=10pt,
    framexleftmargin=5pt,
    breaklines=true
]
# To train the model(s) on EMNIST, run this command:
python main.py --train --dataset emnist --model_backbone mobilenet --grouping_volatility --debug
    \end{lstlisting}
    \end{enumerate}
\begin{enumerate}
    \item[(4)] Instructions on how to conduct inference by providing the command line argument used to execute the evaluation script. The example shown in Listing \ref{code:testInstructions} explains the command used to execute the testing script for TRUNK. It demonstrates the additional arguments required to conduct inferences on the network such as the type of dataset being used, and what deep neural network design (MobileNet \cite{howard_mobilenets_2017} or VGG \cite{simonyan_very_2014}) we are using. 

\begin{lstlisting}[
    language=TeX,
    caption=Example of providing instructions on how to execute the testing for the TRUNK network,
    label={code:testInstructions},
    basicstyle=\ttfamily\small,
    frame=single,
    xleftmargin=10pt,
    framexleftmargin=5pt,
    breaklines=true
]
# To evaluate the model on EMNIST, run:
python main.py --infer --dataset emnist --model_backbone mobilenet --grouping_volatility 
    \end{lstlisting}
    \end{enumerate}
\begin{enumerate}
    \item[(5)] Links to the pre-trained weights of the network as shown in Figure \ref{fig:readmeDocumentation}.
    \item[(6)] Summary of the results, tabulated as shown in Figure \ref{fig:readmeDocumentation}.
        \begin{figure}[h]
        \centering
        \includegraphics[width=0.5\textwidth]{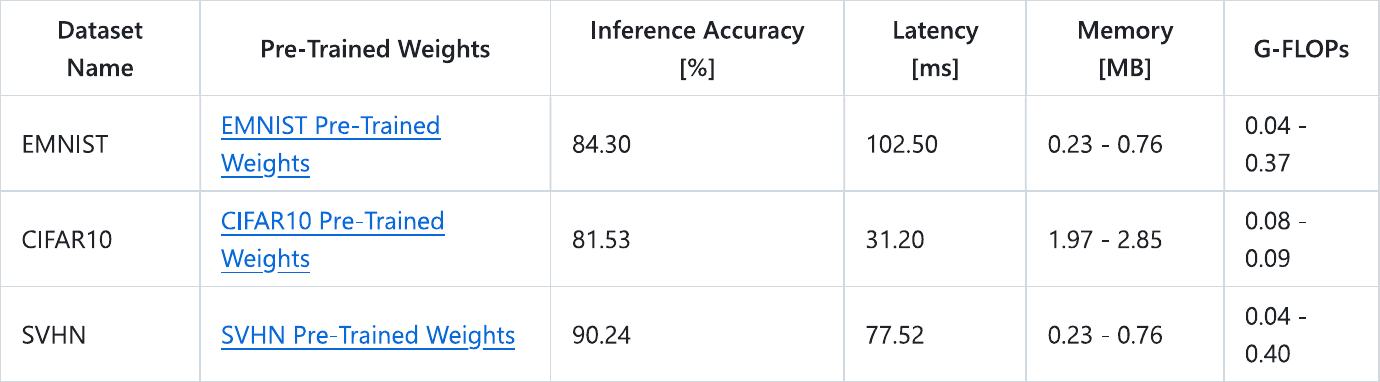}
        \caption{Documenting the links to the Pre-Trained Weights and a summary of their results} 
        \label{fig:readmeDocumentation}
    \end{figure}
\end{enumerate}

Stojnic \cite{johnson2020datascience} uncovered a significant correlation between the number of GitHub stars a repository garnered and the organization of its README file, particularly if it included the previously mentioned details. In fact Meta's DinoV2 \cite{dinov2} followed this documentation structure and achieved nearly 7.7k GitHub stars. This finding \cite{johnson2020datascience} suggests that such repositories are perceived by peers as being of high quality and having reproducible results. As a result, we modify the README file provided for the TRUNK software to follow this structure as well. 


\section{Discussion} \label{sec:guidelines}
Through our experiments, we observed and demonstrated that the current guidelines for improving the reproducibility of deep learning software are integral to enhancing research transparency. These guidelines disclose the methodologies used by the authors to achieve the reported results and serve as a blueprint for fellow researchers to reproduce. We expand on these existing guidelines by advocating the additional guidelines:
\begin{enumerate}
    \item[(1)] To facilitate the replication of the original software environment and dependencies, document all primary GPU-compatible dependencies in the manifest
    \item[(2)] Perform a sensitivity analysis to reveal trends in the DL model's behavior, helping researchers adjust for non-determinism and closely reproduce the original implementation.
\end{enumerate}

\begin{tcolorbox} [width=\linewidth, top=1pt, bottom=1pt, left=2pt, right=2pt, title=Compilation of current guidelines with our extensions (in blue) for improving the reproducibility of DL software]
    \begin{itemize}
        \item \textcolor{blue}{To set up the software environment, provide a manifest compatible with either \texttt{pip} \cite{pip} or \texttt{conda} \cite{conda}, listing only the primary dependencies that support GPU functionality}
        \item Document the hardware used \cite{ml_chem, pineau2020}
        \item Initialize a seed \cite{chen2022, semmelrock2023reproducibility, seed} to limit the randomness in the software
        \item Disclose the data processing and training regime used \cite{ml_chem, natureTransp, torchDistill}
        \item \textcolor{blue}{Conduct a sensitivity analysis to enable fellow researchers to understand the behavior of the DL network across a wide spectrum of parameters}
        \item Release the source code that would reproduce the paper onto public platforms \cite{ml_chem, natureTransp, outofbox, pineau2020, semmelrock2023reproducibility}
        \item Instructions on how to execute the source code \cite{johnson2020datascience}
        \item Access to the Pre-Trained Weights \cite{johnson2020datascience}
        \item Need for proper documentation alongside the DL software \cite{ml_chem, pineau2020, Gundersen_Kjensmo_2018, johnson2020datascience}
    \end{itemize}
\end{tcolorbox}
\section{Threats to Validity}
\paragraph{Construct} We found that providing datasets and source code alone is insufficient for reproducing experimental results in deep learning. In addition to clear documentation of the software environment and training pipeline, the inherent stochasticity of these models necessitates thorough sensitivity analysis. To demonstrate this, we re-trained TRUNK with identical pipelines, random seed, and GPU hardware, yet obtained different tree structures (Figure \ref{fig:reproResults}). The sensitivity analysis enabled us to identify and adjust the relevant parameter, resulting in a comparable structure and accuracy. This highlights sensitivity analysis as a crucial element of reproducible deep learning research.

\paragraph{Internal} In this study, there was a potential risk of overlooking a critical aspect of reproducibility. However, by employing a meticulous and systematic approach, we addressed this risk with steps that began by replicating the software environment and progressed to focusing on the training implementation of our chosen case study.\newline

\paragraph{External} For \textbf{RQ1} and \textbf{RQ2}, the selected case study, TRUNK, may not represent all DL algorithms. While the guidelines devised might enhance TRUNK's reproducibility, they could lack guidelines specific to other DL methods. In other words, the case study approach may not be fully generalizable across diverse DL methods. However, the complexity of our case study, which involves a hierarchical neural network composed of multiple smaller DNNs, provides a substantial degree of generalizability to address our research questions effectively.\newline

\section{Conclusion}
Given the stochastic nature of deep learning models, ensuring the reliability, verifiability, and applicability of their findings across various environments is crucial. Therefore, reproducibility in deep learning is essential. As a result, this paper extends the current set of guidelines to enhance the reproducibility of deep learning software.

We used the Tree-Based Unidirectional Neural Network (TRUNK) as a case study to assess its reproducibility. We began by assessing the original authors' documented procedures for replicating their software environment and made improvements where necessary. When we attempted to reproduce the training of the TRUNK network on a specific dataset, we observed nearly a 25\% discrepancy in model accuracy. To address this, we conducted a sensitivity analysis to understand the performance of the model under various conditions. This analysis helped us identify and correct discrepancies in the TRUNK architecture, despite using the same training recipe, thus circumventing the challenges of inherent non-determinism. 

Furthermore, we examined the impact of documenting the training and data-processing pipeline. By experimenting with various training recipes, we achieved nearly a 14\% increase in accuracy. These findings underscore the critical importance of fully disclosing the training and data processing pipelines, which were initially absent. Together, the guidelines outlined in this paper should be the recommended practices for improving the reproducibility of deep learning software. 

\section{Acknowledgements}
This work is supported by Cisco, Google, and the US National Science Foundation (NSF) under awards \#2107230 and \#2107020.
We acknowledge using ChatGPT for grammar and structural improvements while affirming all ideas and findings are our own original research.

\clearpage

 \Urlmuskip=0mu plus 1mu\relax
 \sloppy
 \bibliographystyle{elsarticle-num}

\clearpage
\end{document}